\documentclass{research}

\usepackage{amsmath,amsfonts,bm}

\def\eqref#1{equation~\ref{#1}}

\def\1{\bm{1}}

\def\vzero{{\bm{0}}}

\def\vg{{\bm{g}}}
\def\vh{{\bm{h}}}

\def\vu{{\bm{u}}}
\def\vv{{\bm{v}}}

\def\vx{{\bm{x}}}
\def\vy{{\bm{y}}}
\def\vz{{\bm{z}}}

\def\mJ{{\bm{J}}}

\DeclareMathAlphabet{\mathsfit}{\encodingdefault}{\sfdefault}{m}{sl}
\SetMathAlphabet{\mathsfit}{bold}{\encodingdefault}{\sfdefault}{bx}{n}

\newcommand{\Ls}{\mathcal{L}}
\newcommand{\R}{\mathbb{R}}

\usepackage{hyperref}
\usepackage{url}

\usepackage{amsmath}
\usepackage{amssymb}

\usepackage{etoc}
\usepackage{titletoc}

\usepackage{hyperref}

\usepackage{booktabs}
\usepackage[table]{xcolor}
\usepackage{multirow}

\usepackage{array}

\usepackage{amsthm}

\usepackage{enumitem}

\usepackage{bm}
\usepackage{tcolorbox}
\tcbuselibrary{skins}
\usepackage{cleveref}

\Crefname{equation}{Equation}{Equations}
\crefname{equation}{Eq.}{Eqs.}
\Crefname{figure}{Figure}{Figures}
\crefname{figure}{Fig.}{Figs.}
\Crefname{table}{Table}{Tables}
\crefname{table}{Table}{Tables}
\Crefname{section}{Section}{Sections}
\crefname{section}{Section}{Sections}
\Crefname{appendix}{Appendix}{Appendices}
\crefname{appendix}{Appendix}{Appendices}
\AddToHook{cmd/appendix/before}{%
    \crefalias{section}{appendix}%
    \crefalias{subsection}{appendix}
}

\definecolor{sand}{RGB}{248,245,235}  
\definecolor{sandborder}{RGB}{220,215,200}
\definecolor{darkblue}{RGB}{40,70,140}   
\definecolor{parchment}{RGB}{252,249,240}
\definecolor{parchmentborder}{RGB}{222,215,195}
\definecolor{sage}{RGB}{244,248,244}
\definecolor{sageborder}{RGB}{200,210,200}
\definecolor{lavender}{RGB}{247,246,250}
\definecolor{lavenderborder}{RGB}{210,205,225}
\definecolor{riskbg}{RGB}{255,246,246}
\definecolor{riskborder}{RGB}{205,105,105}

\definecolor{LyapuCol}{RGB}{235,241,247}   
\definecolor{Satisfied}{RGB}{226,239,218}  
\definecolor{Violated}{RGB}{249,220,220}   
\definecolor{efficiencyred}{RGB}{187,94,96}
\definecolor{Gray}{gray}{0.9}

\newcolumntype{Y}{>{\columncolor{LyapuCol}}r}

\newcommand{\best}[1]{\textbf{#1}}
\newcommand{\second}[1]{\underline{#1}}

\newcommand{\vio}[1]{\cellcolor{Violated}#1}

\newcommand{\viooff}[1]{#1}
\newcommand{\satoff}[1]{0}

\title{LyapuFlow: Controlling Generative Flows with Lyapunov Feedback for Inverse Problems}

\author[1,4]{Minseon Gwak}
\author[1,2]{Hans Hao-Hsun Hsu}
\author[3, *]{\\Danielle C. Maddix}
\author[1,4]{N. Benjamin Erichson}

\affiliation[1]{Lawrence Berkeley National Lab}
\affiliation[2]{Georgia Institute of Technology}
\affiliation[3]{Siemens}
\affiliation[4]{International Computer Science Institute}

\correspondence{Minseon Gwak @ \email{mgwak@lbl.gov}; and N. Benjamin Erichson @ \email{erichson@lbl.gov}.}

\abstract{Pretrained flow models are now widely used as generative priors in science and vision, where inference-time guidance enables test-time constraints without retraining. Existing methods use projection, posterior sampling, or iterative optimization of the generative trajectory. We propose LyapuFlow, an alternative based on Lyapunov feedback control. At each sampling step, LyapuFlow predicts the terminal sample towards which the current flow is evolving, and evaluates the constraint violation on this prediction. Then, we compute the minimum-norm control that satisfies a prescribed Lyapunov decrease condition. The resulting control remains inactive when the uncontrolled dynamics already reduce the constraint violation at the prescribed rate. Otherwise, it provides a corrective update within a feedback trust region that prevents the control from dominating the pretrained dynamics. We demonstrate LyapuFlow in both data and latent spaces, outperforming alternatives spanning different mechanisms for test-time constraint enforcement in scientific machine learning and image inverse problems.}

\begin{document}

\maketitle

\renewcommand{\thefootnote}{*}
\footnotetext{Work done while at AWS.}
\renewcommand{\thefootnote}{\arabic{footnote}} 

\section{Introduction}

Flow matching has emerged as a flexible framework for modeling complex data distributions~\citep{lipman2023flow}. In many downstream applications, the quality of generated samples critically depends on whether they satisfy task-specific constraints. In scientific domains, for example, generated solutions to partial differential equations (PDEs) need to be consistent with hard physical requirements, including initial condition and conservation law. In image inverse problems, generated samples must reproduce measurements degraded by a forward operator. Conditional generative models can incorporate constraints available during training~\citep{tong2024improving, keegan2026manifold}, but constraints specified at inference time motivate \emph{test-time constrained generation}, where the sampling trajectory of a pretrained model is steered to satisfy the specified constraints.

Existing test-time constrained generation methods modify the sampling trajectory through iterative optimization or online guidance~\citep{daras2024survey}. Iterative trajectory optimization repeatedly integrates the generative dynamics to evaluate outcomes and refine the trajectory~\citep{DFlow, FlowGrad, OCFlow}, whereas online guidance applies instantaneous corrections during a single sampling rollout. Specifically, projection-based methods map intermediate states or predicted endpoints toward a constraint-defined feasible set~\citep{PDM, ECI, PCFM}. These projections strongly enforce feasibility, but can overly prioritize constraint satisfaction over sample fidelity. Posterior guidance methods bias the trajectory toward samples with higher posterior probability of satisfying the constraints~\citep{DPS, FlowChef, FlowDPS}, typically relying on task- and model-dependent guidance strengths. Control-theoretic approaches for generative policies in robotics~\citep{UniConFlow, SADFlower} construct feedback directly from constraints on the evolving state, leaving feedback based on explicit terminal predictions unexplored.

\begin{figure}[t!]
    \centering
    \includegraphics[width=0.9\linewidth]{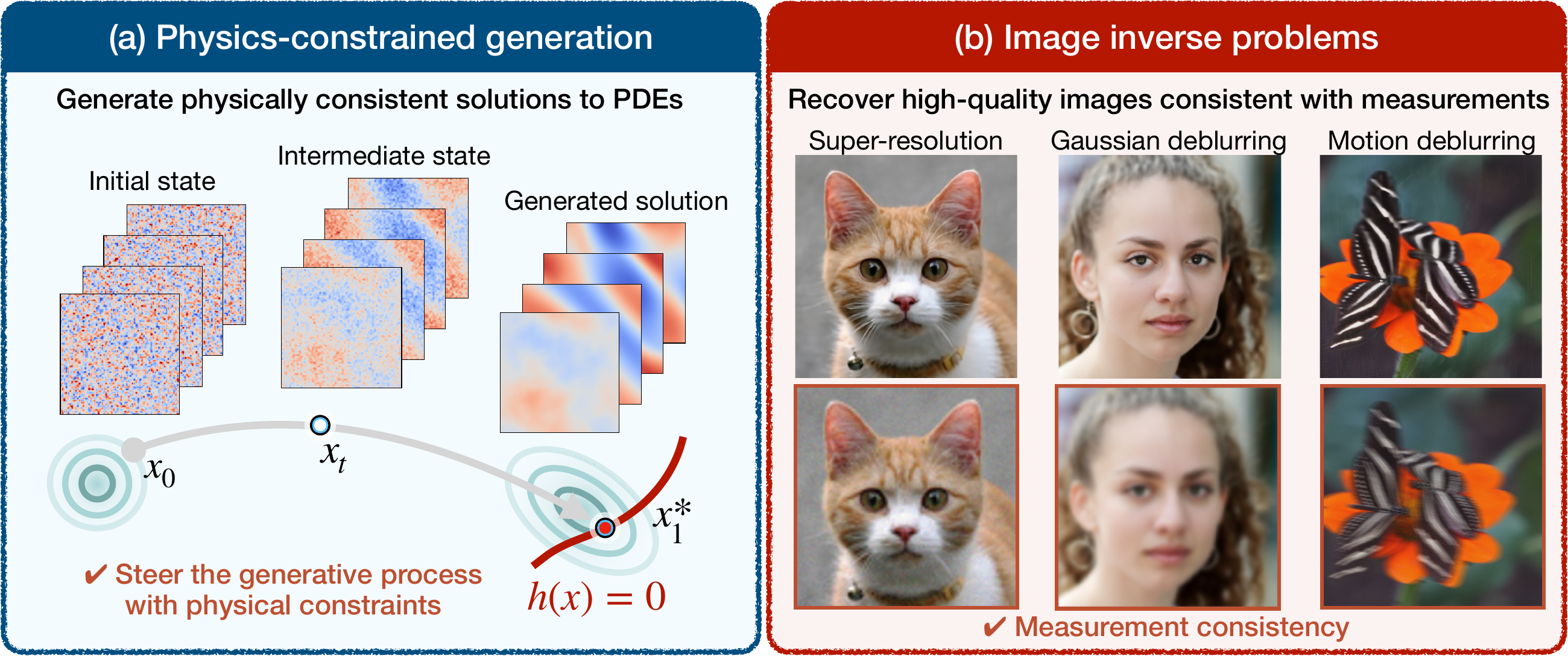}
    \vspace{-0.5em}
    \caption{Test-time constraints for scientific and image inverse problems. (a) Physical constraints guide PDE solution generation. (b) Measurements guide image super-resolution and deblurring.}
    \label{fig:intro}
\end{figure}

In this work, we propose an alternative guidance using Lyapunov feedback control~\citep{khalil2002nonlinear} on constraint satisfaction of predicted terminal samples. Given a pretrained flow $\vv_\theta(\vx_t,t)$ transporting $\vx_0\sim p_0$ to $\vx_1\sim p_1$, we augment its dynamics with an inference-time control $\vu(\vx_t,t)$:
\begin{tcolorbox}[
  colback=parchment,
  colframe=sandborder,
  boxrule=0.3pt,
  left=8pt,
  right=8pt,
  top=1pt,
  bottom=1pt
]
\begin{equation}\label{eq:controlled_flow}
\dot{\vx}_t
=
\vv_\theta(\vx_t,t)
+
\vu(\vx_t,t),
\qquad
\text{with test-time constraint}
\qquad
\vh(\vx_1)=\vzero,
\end{equation}
\end{tcolorbox}
where $p_0$ and $p_1$ denote the base and target distributions.
The difficulty is that the control acts at intermediate states $\vx_t$, while the constraint is defined on the terminal sample $\vx_1$. We propose LyapuFlow to address this by predicting the terminal sample toward which the current flow is evolving and evaluating the constraint on this
prediction. The resulting constraint energy defines a terminal-aware Lyapunov function. At each integration step, LyapuFlow computes a closed-form minimum-norm feedback from a prescribed Lyapunov decrease condition. The feedback vanishes when the pretrained flow already satisfies this condition, allowing the sample to follow the original generative trajectory. Otherwise, it provides a corrective update within a flow-relative trust region that prevents excessive intervention, without iterative optimization.
Our main contributions are as follows: 
\begin{itemize}[leftmargin=*]
    \item We introduce LyapuFlow, a test-time feedback controller for pretrained flow models. LyapuFlow predicts the terminal sample from each intermediate state and computes a closed-form minimum-norm feedback from a prescribed Lyapunov decrease condition (see \cref{sec:method}).

    \item We characterize how the continuous-time feedback behaves under finite-step integration and introduce a feedback trust region to limit overly large corrections (see \cref{sec:discrete_analysis}). We further extend LyapuFlow to latent-space models by propagating constraint gradients through the decoder and optionally restoring the velocity-Jacobian contribution for finer-scale refinement (see \cref{sec:latent}).

    \item We evaluate LyapuFlow on physics-constrained generation in data space (see \cref{sec:experiment-pde}) and image inverse problems in latent space (see \cref{sec:experiment-image}). Across both settings, LyapuFlow improves sample fidelity while maintaining competitive constraint satisfaction.
\end{itemize}

\section{Background and Related Work}
\label{sec:background}

In this section, we review existing approaches for enforcing a general equality constraint $\vh(\vx)=0$. 
As illustrated in \cref{fig:intro}, $\vh$ varies across domains, ranging from physical constraints in PDEs to measurement-consistency constraints in image inverse problems. 
Below, we examine how existing methods for steering generative trajectories toward these constraints differ in their mechanisms.

\textbf{Projection-based enforcement.}
Projection methods explicitly correct generated samples toward the constraint set \citep{probconserv, PDM}. For a differentiable residual $\vh:\mathbb{R}^d\to\mathbb{R}^m$, consider the local linearization of the constraint $\vh(\vx)+\mJ_{\vh}(\vx)\Delta\vx=\vzero$, where $\mJ_{\vh}=\partial\vh/\partial\vx$. Assuming $\mJ_{\vh}$ has full row rank (typically with $d>m$), the minimum-norm correction is
\begin{equation}
\vx^{+}
=
\vx
-
\mJ_{\vh}(\vx)^\top
\left[
\mJ_{\vh}(\vx)\mJ_{\vh}(\vx)^\top
\right]^{-1}
\vh(\vx).
\label{eq:projection_background}
\end{equation}
ECI sampling applies constraint corrections to an extrapolated endpoint using a problem-specific correction operator and interpolates the corrected endpoint with $x_0$ for the next sampling step~\citep{ECI}. PCFM extends projection-based correction to differentiable nonlinear equality constraints using Jacobian-based Gauss--Newton updates~\citep{probharde2e, PCFM}. For nonlinear constraints, \cref{eq:projection_background} requires solving an $m\times m$ linear system at each projection, which severely limits scalability.

\textbf{Posterior and gradient-based guidance.}
Another method class guides generation using likelihood or task-objective gradients. For a measurement $\vy\in\mathbb{R}^m$, posterior sampling targets
\begin{equation}
p(\vx_1\mid \vy)\propto p_1(\vx_1)p(\vy\mid\vx_1),
\label{eq:posterior_guidance}
\end{equation}
where $p_1$ denotes the pretrained generative distribution and $p(\vy\mid\vx_1)$ denotes the measurement likelihood. DPS approximates sampling from \cref{eq:posterior_guidance} by estimating the measurement-likelihood gradient to guide the reverse diffusion process~\citep{DPS}. FlowChef applies efficient task-objective guidance to rectified flow models, while avoiding backpropagation through the velocity model~\citep{FlowChef}. FlowDPS extends posterior guidance to flow models using measurement-likelihood gradients at estimated clean states, likewise bypassing the velocity-model Jacobian~\citep{FlowDPS}.  ProFlow similarly refines terminal estimates to improve physical consistency in PDE problems~\citep{ProFlow}. These methods steer the trajectory without explicit projection onto the constraint set, but suitable guidance scales can vary widely across tasks.

\textbf{Control and trajectory optimization.}
Control theory provides another approach to modifying pretrained generative dynamics. FlowGrad and OC-Flow iteratively refine a complete sequence of control inputs to improve the terminal objective~\citep{FlowGrad,OCFlow}. OC-Flow formulates this refinement as an optimal control problem and solves it using Pontryagin's Maximum Principle~\citep{pontryagin1987mathematical}, balancing terminal loss and control cost:
\begin{equation}
\vu^*_{[0,1]}
\in \arg\min_{\vu_{[0,1]}}
\mathcal L(\vx_1)+\lambda\int_0^1\|\vu_t\|^2\,dt,
\end{equation}
where $\lambda>0$ weights the control cost.
SAD-Flower and UniConFlow compute instantaneous $\vu_t$ through quadratic programs based on Lyapunov and barrier conditions for robotic planning~\citep{SADFlower,UniConFlow}. These conditions measure constraint violations on partially denoised samples. Preserving fidelity to the generative prior remains a challenge.

\section{LyapuFlow: Lyapunov Feedback for Constrained Generation}
\label{sec:method}

\begin{figure}[!t]
    \centering
    \includegraphics[width=0.9\linewidth]{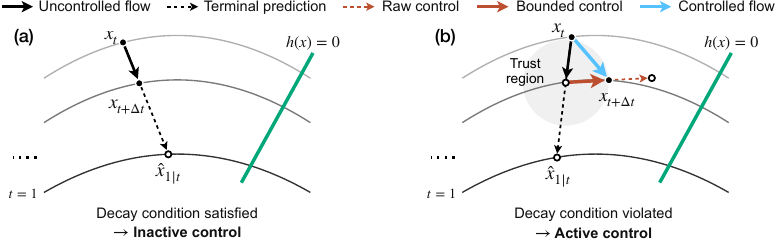}
    \vspace{-0.5em}
    \caption{Illustration of LyapuFlow. At each integration step, LyapuFlow evaluates the pretrained velocity, predicts the terminal sample, computes the minimum-norm feedback from the Lyapunov decrease condition, bounds the feedback using the trust region, and advances the controlled flow. (a) When the pretrained dynamics already satisfy the Lyapunov decrease condition, no feedback is applied. (b) Otherwise, LyapuFlow applies the bounded control.}
    \label{fig:method}
\end{figure}

In this section, we develop LyapuFlow for the controlled generative dynamics introduced in \cref{eq:controlled_flow}. Given a pretrained velocity field $\vv_\theta$ and a terminal constraint $\vh(\vx_1)=\vzero$, our goal is to construct an inference-time feedback control $\vu(\vx_t,t)$ that reduces terminal constraint violation while avoiding excessively large corrections to the generative dynamics. The main difficulty is that the control acts on intermediate states $\vx_t$, whereas the constraint is defined only on the terminal sample $\vx_1$. \Cref{fig:method} provides an overview of the LyapuFlow sampling, illustrating how terminal constraints are propagated to intermediate states while the feedback remains tied to the pretrained generative dynamics.

The control is applied at intermediate states $\vx_t$, whereas the constraint
is defined on the terminal sample $\vx_1$. To obtain a quantity that can
drive local feedback, we predict the terminal sample $\hat{\vx}_{1|t}$ using a
first-order extrapolation over the remaining integration interval
\begin{equation}
\hat{\vx}_{1|t}
=
\vx_t
+
(1-t)\vv_\theta(\vx_t,t).
\label{eq:terminal_prediction}
\end{equation}
To quantify violation of the terminal constraint, we use the quadratic
residual loss
$\Ls(\vx)=\frac{1}{2}\|\vh(\vx)\|_2^2$, which is nonnegative and vanishes
when the constraint is satisfied. Evaluating this loss at the predicted
terminal sample defines
\begin{equation}
V(\vx_t,t)
=
\Ls\left(\hat{\vx}_{1|t}\right)
=
\frac{1}{2}
\left\|
\vh\left(\hat{\vx}_{1|t}\right)
\right\|_2^2.
\label{eq:lyapunov_function}
\end{equation}
Thus, $V$ can be interpreted as the constraint energy of the predicted
terminal sample. It provides a terminal-aware quantity at each intermediate
state, while $\hat{\vx}_{1|t}=\vx_1$ ensures that
$V(\vx_1,1)=\Ls(\vx_1)$ at the end of generation.
We use $V$ as a Lyapunov function to determine the feedback control. Along
the controlled dynamics
$\dot{\vx}_t=\vv_\theta(\vx_t,t)+\vu_t$, its total derivative is
\begin{equation}
\dot{V}
=
\partial_t V
+
\nabla_{\vx_t}V^\top \vv_\theta(\vx_t,t)
+
\nabla_{\vx_t}V^\top \vu_t .
\label{eq:V_derivative}
\end{equation}
We choose the control so that the predicted terminal constraint violation decreases as
\begin{equation}
\dot{V}
\leq
-\alpha V,
\qquad
\alpha \geq 0,
\label{eq:lyapunov_condition}
\end{equation}
where $\alpha$ specifies the desired rate of decrease. Substituting the controlled dynamics into~\cref{eq:lyapunov_condition} gives a linear constraint on the instantaneous control
\begin{equation}
\nabla_{\vx_t}V^\top \vu_t
\leq
-
\partial_t V
-
\nabla_{\vx_t}V^\top \vv_\theta(\vx_t,t)
-
\alpha V.
\label{eq:control_constraint}
\end{equation}
Following the classical pointwise minimum-norm construction in~\citet{freeman2008robust}, we select the smallest instantaneous correction that satisfies this condition
\begin{equation}
\begin{aligned}
\mathrm{minimize}_{\vu_t}\;\;
\frac{1}{2}\|\vu_t\|_2^2 \qquad
\text{s.t.}\;\;
\nabla_{\vx_t}V^\top \vu_t
+
\partial_t V
+
\nabla_{\vx_t}V^\top \vv_\theta(\vx_t,t)
+
\alpha V
\leq 0 .
\end{aligned}
\label{eq:min_norm_control}
\end{equation}
The constraint in \cref{eq:min_norm_control} defines a half-space in the control space, with minimum-norm solution (see \cref{app:kkt-derivation} for derivation).
\begin{tcolorbox}[
  colback=parchment,
  colframe=sandborder,
  boxrule=0.3pt,
  left=8pt,
  right=8pt,
  top=1pt,
  bottom=1pt
]
\begin{equation}
\vu_t^\star
=
-\eta_t\nabla_{\vx_t}V,
\qquad
\eta_t
=
\frac{
\left[
\partial_t V
+
\nabla_{\vx_t}V^\top \vv_\theta(\vx_t,t)
+
\alpha V
\right]_+
}{
\|\nabla_{\vx_t}V\|_2^2
}.
\label{eq:lyapuflow_control}
\end{equation}
\end{tcolorbox}
The scalar $\eta_t$ is an adaptive feedback gain determined by the current
state of the generative flow. When the term inside $[\cdot]_+$ is nonpositive,
the pretrained flow already reduces $V$ at the requested rate and
$\eta_t=0$. Otherwise, $\eta_t$ is the smallest gain along
$-\nabla_{\vx_t}V$ that satisfies the Lyapunov decrease condition. Unlike
fixed-scale gradient guidance, the strength of the correction therefore
adapts to the nominal flow, the current constraint violation, and the local
gradient of $V$.

\textbf{Flow-aware feedback.}
The gradient $\nabla_{\vx_t}V$ depends on both the terminal constraint and the local sensitivity of the pretrained flow. Differentiating \cref{eq:lyapunov_function} through the terminal prediction gives
\begin{equation}
\nabla_{\vx_t}V
=
\nabla_{\hat{\vx}_{1|t}}\Ls
+
(1-t)\mJ_{\vv}(\vx_t,t)^\top
\nabla_{\hat{\vx}_{1|t}}\Ls,
\label{eq:flow_aware_gradient}
\end{equation}
where $\mJ_{\vv}(\vx_t,t) = \frac{\partial \vv_\theta(\vx_t,t)}{\partial \vx_t}$.
The first term on the right-hand side captures the direct effect of perturbing the current state on the terminal prediction. The second term captures the indirect effect through the state dependence of the pretrained velocity field.
The factor $(1-t)$ reduces the influence of $\mJ_{\vv}$ near the terminal time, with $\nabla_{\vx_t}V$ approaching $\nabla_{\hat{\vx}_{1|t}}\Ls$.

Computing the first term requires only the constraint-loss gradient at the terminal prediction. The second term additionally requires differentiating through the pretrained velocity model, increasing the cost of the feedback. We only need to compute the vector--Jacobian product $\mJ_{\vv}^\top\nabla_{\hat{\vx}_{1|t}}\Ls$ using reverse-mode automatic differentiation~\citep{vjp}. 
In the latent-space extension, we further consider omitting this Jacobian-dependent term for efficiency, as discussed in~\cref{sec:latent}. 
See ~\cref{app:pde-efficiency,app:image-efficiency} for efficiency analyses.

The controller also depends on the explicit time derivative $\partial_tV$. Differentiating the terminal prediction with respect to $t$ and combining the result with \cref{eq:flow_aware_gradient} yields
\begin{equation}
\partial_t V
+
\nabla_{\vx_t}V^\top\vv_\theta
=
(1-t)
\nabla_{\hat{\vx}_{1|t}}\Ls^\top
\left[
\partial_t\vv_\theta
+
\mJ_{\vv}\vv_\theta
\right].
\label{eq:transport_term}
\end{equation}
The term in brackets is the rate of change of the pretrained velocity along its own trajectory. The uncontrolled part of $\dot V$ measures how the predicted terminal constraint changes because the learned velocity field varies along the flow. 

\subsection{Discretization and Feedback Trust Region}
\label{sec:discrete_analysis}

We derive \cref{eq:lyapuflow_control} from a continuous-time Lyapunov decrease condition. The sampler applies this feedback on a discretized time grid $\{t_k\}_{k=0}^{K}$, where $K$ denotes the number of function evaluations (NFEs). 
Let $\vx_k$ denote the state at integration step $k$ and $V_k:=V(\vx_k,t_k)$. Under Euler integration, the dynamics controlled by $\vu_k^\star=-\eta_k\nabla_{\vx_k}V_k$ is
\begin{equation}
\vx_{k+1}
=
\vx_k
+
\Delta t_k\,\vv_\theta(\vx_k,t_k)
-
\Delta t_k\,\eta_k\nabla_{\vx_k}V_k.
\label{eq:discrete_sampler}
\end{equation}

After this discretization, a feedback magnitude chosen to satisfy the instantaneous condition in \cref{eq:lyapunov_condition} can be too large for a finite-step update, moving the sample beyond the region in which the local decrease was evaluated~\citep{taylor2021sampled}. These large corrections are also undesirable for generative flows, where deviations from the learned transport path can degrade sample fidelity~\citep{saini2026rectified}. 
We introduce a feedback trust region and characterize the resulting trade-off.

\begin{figure}[!t]

\centering
\begin{minipage}[t]{0.48\textwidth}
\begin{algorithm}[H]
\small
\caption{LyapuFlow (Data space)}
\label{alg:data}
\begin{algorithmic}[1]
\vspace{-0.2em}
\Require $\vv_\theta$, decay rate $\alpha$,
trust region factor $c_{\mathrm{tr}}$
\State $\vx_0 \sim p_0$
\For{$k=0,\ldots,K-1$}
    \State $\vv_k \gets \vv_\theta(\vx_k,t_k)$
    \State $\hat{\vx}_{1|k}
        \gets \vx_k+(1-t_k)\vv_k$
    \State $V_k \gets \Ls(\hat{\vx}_{1|k})$
    \State $\vg_k \gets \nabla_{\vx_k}V_k,\quad
        q_k \gets \partial_tV(\vx_k,t_k)$
    \State $\vu_k^\star \gets
        -\dfrac{[q_k+\vg_k^\top\vv_k+\alpha V_k]_+}
        {\|\vg_k\|_2^2}\,\vg_k$
    \State $\tilde{\vu}_k \gets
        \min\Big(
        1,\dfrac{c_{\mathrm{tr}}\|\vv_k\|_2}
        {\|\vu_k^\star\|_2}
        \Big)\vu_k^\star$
    \State $\vx_{k+1} \gets
        \vx_k+\Delta t_k(\vv_k+\tilde{\vu}_k)$
\EndFor
\State \Return $\vx_K$
\end{algorithmic}
\end{algorithm}
\end{minipage}
\hfill
\begin{minipage}[t]{0.48\textwidth}
\begin{algorithm}[H]
\scriptsize
\caption{LyapuFlow (Latent space)}
\label{alg:latent}
\begin{algorithmic}[1]
\Require $\vv_\theta$,
$\alpha$, $c_{\mathrm{tr}}$, decoder $\mathcal D$,
gradient switch step $k_{\mathrm{sg}}$
\State $\mathbf z_0 \sim p_0$
\vspace{0.45em}
\For{$k=0,\ldots,K-1$}
\vspace{0.45em}
    \State $\bar{\vv}_k \gets
        \begin{cases}
        \operatorname{stopgrad}\big(\vv_\theta(\mathbf z_k,t_k)\big), & k<k_{\mathrm{sg}},\\
        \vv_\theta(\mathbf z_k,t_k), & \text{otherwise}
        \end{cases}$
    \State $\hat{\mathbf z}_{1|k}
        \gets \mathbf z_k+(1-t_k)\bar{\vv}_k$
    \State $V_k \gets
        \Ls\left(\mathcal D(\hat{\mathbf z}_{1|k})\right)$
    \State $\vg_k \gets \nabla_{\mathbf z_k}V_k,\quad
        q_k \gets \partial_tV(\mathbf z_k,t_k)$
    \State $\vu_k^\star \gets
        -\dfrac{[q_k+\vg_k^\top\bar{\vv}_k+\alpha V_k]_+}
        {\|\vg_k\|_2^2}\,\vg_k$
    \State $\tilde{\vu}_k \gets
        \min\Big(
        1,\dfrac{c_{\mathrm{tr}}\|\bar{\vv}_k\|_2}
        {\|\vu_k^\star\|_2}
        \Big)\vu_k^\star$
    \State $\mathbf z_{k+1} \gets
        \mathbf z_k+\Delta t_k(\bar{\vv}_k+\tilde{\vu}_k)$
\vspace{0.5em}
\EndFor
\vspace{0.5em}
\State \Return $\mathcal D(\mathbf z_K)$
\end{algorithmic}
\end{algorithm}
\end{minipage}
\par\medskip
\begin{minipage}{\linewidth}
\end{minipage}
\end{figure}

Let $\tilde{\vu}_k$ denote the actual control applied at integration step $k$. We quantify the resulting violation of the continuous-time Lyapunov condition by
\begin{equation}
\xi_k
=
\left[
\partial_tV_k
+
\nabla_{\vx_k}V_k^\top
\left(
\vv_k+\tilde{\vu}_k
\right)
+
\alpha V_k
\right]_+ .
\label{eq:discrete_clf_slack}
\end{equation}
If the applied feedback is the closed-form control $\tilde{\vu}_k=\vu_k^\star$ from \cref{eq:lyapuflow_control}, then the Lyapunov condition is satisfied and $\xi_k=0$. When the feedback is \textit{bounded}, $\tilde{\vu}_k$ may differ from $\vu_k^\star$, and $\xi_k$ measures the resulting loss in the requested Lyapunov decrease.

\begin{proposition}[Discrete-time Lyapunov bound]
\label{prop:discrete_stability}
Suppose that $V(\vx,t)$ has an $L_V$-Lipschitz gradient with respect to $(\vx,t)$. Under an Euler step with step size $\Delta t_k$ and applied feedback $\tilde{\vu}_k$,
\begin{equation}
V_{k+1}-V_{k}
\leq
-\alpha\Delta t_k V_k
+
\Delta t_k\xi_k
+
\frac{L_V}{2}\Delta t_k^2
\left(
1+\|\vv_k+\tilde{\vu}_k\|_2^2
\right).
\label{eq:discrete_lyapunov}
\end{equation}
\end{proposition}

The second term on the right-hand side measures the loss in Lyapunov decrease when the feedback is bounded. 
The final term is the discretization error, which grows with the magnitude of the controlled velocity; hence, bounding the control can limit this contribution. 
Therefore, Proposition~\ref{prop:discrete_stability} reveals a trade-off between \textit{preserving the desired Lyapunov decrease} and \textit{limiting the discretization error}, which can be balanced through an appropriate choice of the feedback bound.

To this end, we define the \emph{feedback trust region} to bound the feedback displacement, $\left\| \Delta t_k\,\tilde{\vu}_k \right\|_2 \leq \Delta t_k \bar{u}_k$. 
While explicitly optimizing $\bar{u}_k$ at every step would incur additional computation, we instead use a simple adaptive rule based on the flow magnitude  
\begin{equation} \label{eq:control-trust-region-bound}
    \bar{u}_k=c_{\mathrm{tr}}\|\vv_k\|_2,
\end{equation} 
for a positive constant $c_{\mathrm{tr}}$.
 
%
The resulting control applied at integration step $k$ is
\begin{equation}
\tilde{\vu}_k
=
\frac{\bar{u}_k}
{\max\left(\bar{u}_k,\|\vu_k^\star\|_2\right)}
\vu_k^\star.
\label{eq:control-trust-region}
\end{equation}
\Cref{alg:data} summarizes our sampler. See~\cref{app:discrete_stability} for the proofs and additional analysis.

\subsection{Practical Extension to Latent-Space Models}
\label{sec:latent}

The above formulations directly apply when the flow and constraint are defined in the same data space. In domains with established latent representations, generative models often operate in latent space while constraints remain in data space~\citep{rombach2022high, plsd}. Let $\vz_t$ denote the latent state and $\mathcal D$ the decoder. The terminal prediction then follows the computation path
\begin{equation}
    \vz_t
    \xrightarrow{\;\vv_\theta\;}
    \hat{\vz}_{1|t}
    =
    \vz_t+(1-t)\vv_\theta(\vz_t,t)
    \xrightarrow{\;\mathcal D\;}
    \hat{\vx}_{1|t}
    =
    \mathcal D(\hat{\vz}_{1|t}),
\end{equation}
and we define the Lyapunov function using the decoded terminal prediction $V(\vz_t,t) = \Ls\left(\mathcal D(\hat{\vz}_{1|t})\right)$. Consequently, the constraint gradient with respect to the latent state becomes
\begin{equation}
\begin{aligned}
\nabla_{\vz_t}V
&=
\mJ_{\mathcal D}^\top
\nabla_{\hat{\vx}_{1|t}}\Ls
+
(1-t)\mJ_{\vv}^\top
\mJ_{\mathcal D}^\top
\nabla_{\hat{\vx}_{1|t}}\Ls ,
\end{aligned}
\label{eq:latent_flow_aware_gradient}
\end{equation}
where $\mJ_{\mathcal D}=\partial \mathcal D(\hat{\vz}_{1|t})/\partial \hat{\vz}_{1|t}$ and $\mJ_{\vv}=\partial \vv_\theta(\vz_t,t)/\partial \vz_t$.
This is the latent-space counterpart of \cref{eq:flow_aware_gradient}. The first term propagates the constraint gradient through the decoder alone, whereas the second term accounts for how perturbations of the current latent state alter the predicted endpoint through the velocity field.
In modern latent generative models, the velocity network is often larger than the decoder~\citep{chen2024pixart,fang2024xdit}. 
We therefore consider stopping gradients through the velocity prediction,
$\bar{\vv}_t=\operatorname{stopgrad}(\vv_\theta(\vz_t,t))$,
and use the resulting surrogate gradient 
\begin{equation}
\tilde{\vg}_t
=
\mJ_{\mathcal D}^\top
\nabla_{\hat{\vx}_{1|t}}\Ls.
\end{equation}
The $\operatorname{stopgrad}$ operation removes the dependence of the velocity prediction on both the latent state and time, thus simplifying the effective form of the unbounded feedback to
\begin{equation}
\vu_t^\star
=
-\frac{\alpha V}{\|\tilde{\vg}_t\|_2^2}\tilde{\vg}_t.
\end{equation}
These surrogate derivatives need not coincide with the true derivatives of $V$, so this latent approximation does not in general guarantee decrease of the true Lyapunov function.
In most of our latent-space experiments, this approximation provides sufficient constraint guidance throughout sampling.
For settings requiring high-frequency correction, restoring the $\mJ_\vv$ contribution can be beneficial~\citep{lai2026toward}. Empirically, the decoder-propagated gradient primarily corrects lower-frequency components, while the flow-aware term in \cref{eq:latent_flow_aware_gradient} improves higher-frequency structure. However, including the velocity-Jacobian term from the beginning can overemphasize high-frequency corrections and lead to overly aggressive refinement. Thus, we introduce a gradient switch step $k_{\mathrm{sg}}$ that omits the velocity-
Jacobian contribution at early steps while enabling the full gradient at later steps:
\begin{equation}
\bar{\vv}_k =
\begin{cases}
\operatorname{stopgrad}\left(\vv_\theta(\vz_k,t_k)\right),
& k < k_{\mathrm{sg}},\\
\vv_\theta(\vz_k,t_k),
& k \geq k_{\mathrm{sg}}.
\end{cases}
\label{eq:gradient_switch}
\end{equation}
\Cref{alg:latent} summarizes the LyapuFlow sampling procedure in latent space.

\section{Experimental Results}

We evaluate whether LyapuFlow across two diverse applications: physics-constrained PDE solution generation in data space and image inverse problems in latent space. For comparison, we consider several representative baselines for test-time constrained generation, including DPS~\citep{DPS}, SAD-Flower~\citep{SADFlower}, FlowDPS~\citep{FlowDPS}, ProFlow~\citep{ProFlow}, PCFM~\citep{PCFM}, and OC-Flow~\citep{OCFlow}. 
DPS denotes its flow-based variant, which differentiates through the velocity model.
We also compare to efficient low-iteration (analysis is provided in~\cref{app:image-efficiency}) latent-space-oriented baselines (FlowChef~\citep{FlowChef} and FlowDPS~\citep{FlowDPS}). 
See \cref{app:pde} for experimental details on the PDE experiments and \cref{app:image} for the image experiments, including datasets, pretrained models, constraint definitions, evaluation metrics, baseline configurations, and hyperparameters.

\subsection{Physics-Constrained Generation}
\label{sec:experiment-pde}

\begin{table}[!t]
\centering
\caption{Quantitative evaluation for physics-constrained generation. Bold: best; underline: second-best. All methods apply the same number of final projections (Burgers: 3; other tasks: 1).}
\label{tab:pde-constraint-results}
\scriptsize
\renewcommand{\arraystretch}{0.9}
\resizebox{\textwidth}{!}{%
\begin{tabular}{llYrrrrrrr}
\toprule
\raisebox{-0.5\baselineskip}{Dataset} & \raisebox{-0.5\baselineskip}{Metric \(\downarrow\)} & \raisebox{-0.5\baselineskip}{LyapuFlow} & \raisebox{-0.5\baselineskip}{DPS} & \raisebox{-0.5\baselineskip}{SAD-Flower} & \raisebox{-0.5\baselineskip}{FlowDPS} & \raisebox{-0.5\baselineskip}{ProFlow} & \raisebox{-0.5\baselineskip}{PCFM} & \raisebox{-0.5\baselineskip}{OC-Flow} & \shortstack{FFM\\[-2pt]w/ Proj.} \\
\midrule
\multirow{5}{*}{1D Heat} & \(\mathrm{MMSE}/10^{-2}\) & \best{0.012} & 0.221 & 3.463 & 2.127 & 0.485 & \second{0.115} & 0.134 & 5.206 \\
 & \(\mathrm{SMSE}/10^{-2}\) & \second{0.004} & \best{0.002} & 1.835 & 1.234 & 0.335 & 0.700 & 0.339 & 2.809 \\
 & FPD & \best{0.006} & \second{0.093} & 1.064 & 0.826 & 0.377 & 0.639 & 0.353 & 1.481 \\
 & \(\mathrm{CE(IC)}/10^{-2}\) & 0 & 0 & 0 & 0 & 0 & 0 & 0 & 0 \\
 & \(\mathrm{CE(CL)}/10^{-2}\) & 0 & 0 & 0.003 & 0.001 & 0.002 & 0 & 0.001 & 0.001 \\
\midrule
\multirow{6}{*}{1D Burgers} & \(\mathrm{CMMSE}/10^{-2}\) & \best{0.163} & \second{0.254} & 6.874 & 7.578 & 9.326 & 4.303 & 4.172 & 8.266 \\
 & \(\mathrm{CSMSE}/10^{-2}\) & \best{0.212} & \second{0.377} & 5.708 & 6.234 & 6.896 & 2.171 & 6.150 & 6.715 \\
 & FPD & \best{0.204} & \second{0.649} & 3.967 & 2.911 & 3.307 & 7.072 & 1.020 & 1.414 \\
 & \(\mathrm{CE(IC)}/10^{-2}\) & 0 & 0 & 0.002 & 0.003 & 0.003 & 0 & 0.002 & 0.003 \\
 & \(\mathrm{CE(ND)}/10^{-2}\) & 0.010 & 0.015 & 10 & 18 & 27 & 0 & 13 & 20 \\
 & \(\mathrm{CE(CL)}/10^{-2}\) & 0.005 & 0.007 & 21 & 116 & 152 & 0 & 82 & 124 \\
\midrule
\multirow{5}{*}{1D RD} & \(\mathrm{CMMSE}/10^{-2}\) & \best{0.019} & 0.090 & 3.379 & 2.070 & 1.086 & 0.386 & \second{0.083} & 3.931 \\
 & \(\mathrm{CSMSE}/10^{-2}\) & \best{0.028} & \second{0.053} & 1.903 & 1.138 & 0.819 & 1.225 & 0.194 & 2.196 \\
 & FPD & \best{0.002} & \second{0.008} & 12 & 24 & 4.613 & 3.712 & 1.928 & 7.808 \\
 & \(\mathrm{CE(IC)}/10^{-2}\) & 0 & 0 & 0 & 0 & 0 & 0 & 0 & 0 \\
 & \(\mathrm{CE(NCL)}/10^{-2}\) & 0 & 0 & 0 & 0.002 & 0.001 & 0 & 0 & 0.008 \\
\midrule
\multirow{5}{*}{2D NS} & \(\mathrm{CMMSE}/10^{-2}\) & \best{5.457} & \second{5.871} & 23 & 22 & 22 & 23 & 13 & 23 \\
 & \(\mathrm{CSMSE}/10^{-2}\) & \best{2.480} & \second{2.834} & 6.193 & 6.131 & 6.227 & 13 & 5.876 & 6.270 \\
 & FPD & 0.568 & 0.763 & 0.663 & 0.638 & 0.366 & \second{0.314} & \best{0.263} & 0.653 \\
 & \(\mathrm{CE(IC)}/10^{-2}\) & 0 & 0 & 0 & 0 & 0 & 0 & 0 & 0 \\
 & \(\mathrm{CE(CL)}/10^{-2}\) & 0 & 0 & 0 & 0 & 0 & 0 & 0 & 0 \\
\bottomrule
\end{tabular}
}
\end{table}


\begin{figure}[t!]
    \centering
    \includegraphics[page=9, width=\linewidth]{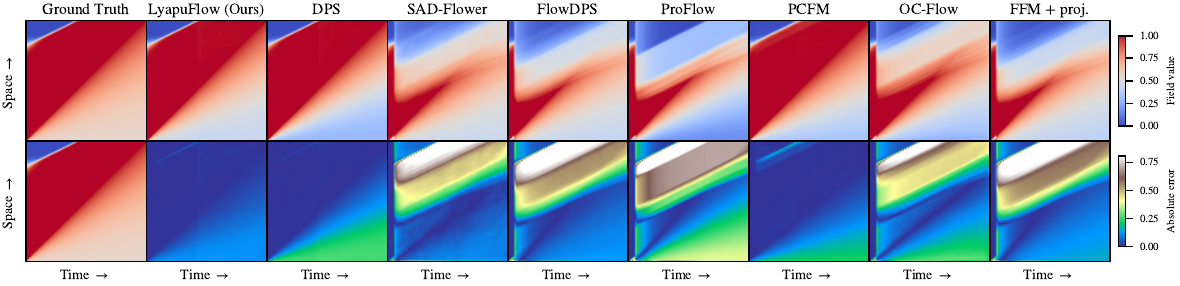}
    \caption{Average 1D Burgers solutions within a fixed-initial-condition group (top) and pointwise absolute error to the ground truth (bottom), subject to the IC, ND constraints over the first five time steps, and CL constraints. See~\cref{app:pde-vis} for results on other PDE datasets.}
    \label{fig:pde-burgers}
\end{figure}

We first consider PDE solution generation, where a pretrained flow is guided to satisfy hard physical requirements not explicitly imposed during training. In this setting, high likelihood under the learned data distribution does not by itself guarantee physical validity. Following ECI~\citep{ECI} and PCFM~\citep{PCFM}, we test on four PDE benchmarks: 1D Heat, 1D Burgers, 1D Reaction--Diffusion (RD), and 2D Navier--Stokes (NS). The benchmarks span smooth diffusive dynamics to evolution with sharp discontinuities. The evaluated constraints include initial conditions (IC), linear conservation laws (CL), nonlinear conservation laws (NCL), and nonlinear dynamics (ND) based on the Godunov update. Performance is assessed in terms of both constraint error (CE) and fidelity of the resulting conditional solution distribution (See \cref{app:pde-metrics} for metrics.).
%

\Cref{tab:pde-constraint-results} shows that LyapuFlow achieves the best MMSE/CMMSE across all four tasks and the best SMSE/CSMSE and FPD on three tasks. LyapuFlow achieves zero or near-zero CE across all tasks, including Burgers after three final projection steps, while the baselines still exhibit non-negligible constraint error under the same projection budget. The main distinction is in sample fidelity, where LyapuFlow more accurately recovers the distribution associated with the specified physical conditions.  \Cref{fig:pde-burgers} illustrates this behavior for Burgers: among the baselines, only LyapuFlow and DPS recover the localized shock structure induced by the nonlinear dynamics. We provide a detailed comparison with DPS in \cref{sec:ablation}.

\begin{table}[t!]
    \centering
    \scriptsize
    \caption{Quantitative evaluation for image inverse problems. All experiments use Stable Diffusion 3.5-Medium with 28 NFEs. Bold: best; Underline: second best.}
    \label{tab:image-quantitative-nfe28}
    \renewcommand{\arraystretch}{0.98} 
    \resizebox{\linewidth}{!}{
    \begin{tabular}{lcccccccccccc}
    \toprule
    \rowcolor{Gray}
    & \multicolumn{12}{c}{AFHQ 1k (768 $\times$ 768)}\\
     & \multicolumn{4}{c}{Super-resolution x4 (Bicubic)} & \multicolumn{4}{c}{Deblurring (Gauss)} & \multicolumn{4}{c}{Deblurring (Motion)}\\
    \cmidrule(lr){2-5} \cmidrule(lr){6-9} \cmidrule(lr){10-13}
    Method & PSNR$\uparrow$ & SSIM$\uparrow$ & FID$\downarrow$ & LPIPS$\downarrow$ & PSNR$\uparrow$ & SSIM$\uparrow$ & FID$\downarrow$ & LPIPS$\downarrow$ & PSNR$\uparrow$ & SSIM$\uparrow$ & FID$\downarrow$ & LPIPS$\downarrow$ \\
    \midrule
    FlowChef &  
     \underline{29.43} & \underline{0.798} & 27.30 & \underline{0.095} & 25.22 & \underline{0.717} & \underline{40.81} & \underline{0.274} & \underline{26.33} & \underline{0.739} & 40.83 & 0.230 \\
    FlowDPS &
     28.24 & 0.746 & \underline{16.05} & 0.122 & \underline{25.41} & 0.680 & 26.87 & 0.244 & 25.95 & 0.693 & \underline{22.78} & \underline{0.212} \\
    \rowcolor{LyapuCol} LyapuFlow (Ours) &
     \textbf{30.05} & \textbf{0.829} & \textbf{9.68} & \textbf{0.045} & \textbf{26.88} & \textbf{0.723} & \textbf{20.21} & \textbf{0.172} & \textbf{27.03} & \textbf{0.740} & \textbf{20.47} & \textbf{0.166} \\
    \midrule
    \rowcolor{Gray}
    & \multicolumn{12}{c}{FFHQ 1k (768 $\times$ 768)}\\
     & \multicolumn{4}{c}{Super-resolution x4 (Bicubic)} & \multicolumn{4}{c}{Deblurring (Gauss)} & \multicolumn{4}{c}{Deblurring (Motion)}\\
    \cmidrule(lr){2-5} \cmidrule(lr){6-9} \cmidrule(lr){10-13}
    Method & PSNR$\uparrow$ & SSIM$\uparrow$ & FID$\downarrow$ & LPIPS$\downarrow$ & PSNR$\uparrow$ & SSIM$\uparrow$ & FID$\downarrow$ & LPIPS$\downarrow$ & PSNR$\uparrow$ & SSIM$\uparrow$ & FID$\downarrow$ & LPIPS$\downarrow$ \\
    \midrule
    FlowChef & 
     \textbf{30.02} & \underline{0.831} & 67.74 & \underline{0.074} & 25.67 & \underline{0.763} & 108.32 & \underline{0.232} & \underline{26.97} & \underline{0.783} & 94.39 & 0.193 \\
    FlowDPS &  
     28.99 & 0.756 & \textbf{27.36} & \underline{0.098} & \underline{26.03} & 0.716 & \underline{46.25} & \underline{0.210} & 26.74 & 0.723 & \textbf{38.52} & \underline{0.177} \\
    \rowcolor{LyapuCol}  LyapuFlow (Ours) &
     \underline{29.97} & \textbf{0.836} & \underline{27.91} & \textbf{0.046} & \textbf{27.56} & \textbf{0.761} & \textbf{41.83} & \textbf{0.144} & \textbf{27.81} & \textbf{0.778} & \underline{40.92} & \textbf{0.138} \\
    \midrule
    \rowcolor{Gray}
    & \multicolumn{12}{c}{DIV2K 0.8k (768 $\times$ 768)}\\
     & \multicolumn{4}{c}{Super-resolution x4 (Bicubic)} & \multicolumn{4}{c}{Deblurring (Gauss)} & \multicolumn{4}{c}{Deblurring (Motion)}\\
    \cmidrule(lr){2-5} \cmidrule(lr){6-9} \cmidrule(lr){10-13}
    Method & PSNR$\uparrow$ & SSIM$\uparrow$ & FID$\downarrow$ & LPIPS$\downarrow$ & PSNR$\uparrow$ & SSIM$\uparrow$ & FID$\downarrow$ & LPIPS$\downarrow$ & PSNR$\uparrow$ & SSIM$\uparrow$ & FID$\downarrow$ & LPIPS$\downarrow$ \\
    \midrule
    FlowChef &
     \underline{24.47} & \underline{0.671} & 51.53 & \underline{0.111} & 
     \underline{20.72} & \underline{0.533} & \underline{98.34} & \underline{0.343} & 
     \underline{21.77} & \underline{0.575} & \underline{86.97} & \underline{0.285} \\
    FlowDPS &
     23.74 & 0.612 & \underline{33.68} & 0.117 & 
     20.55 & 0.482 & 95.47 & 0.346 & 
     21.35 & 0.519 & 76.59 & 0.281 \\
     \rowcolor{LyapuCol} LyapuFlow (Ours) & 
     \textbf{24.78} & \textbf{0.710} & \textbf{25.54} & \textbf{0.051} & \textbf{21.79} & \textbf{0.547} & \textbf{75.67} & \textbf{0.252} & \textbf{22.28} & \textbf{0.589} & \textbf{68.05} & \textbf{0.230} \\
    \bottomrule
    \end{tabular}
    }
\end{table}

\begin{figure}[t]
  \centering
  \includegraphics[width=0.95\textwidth]{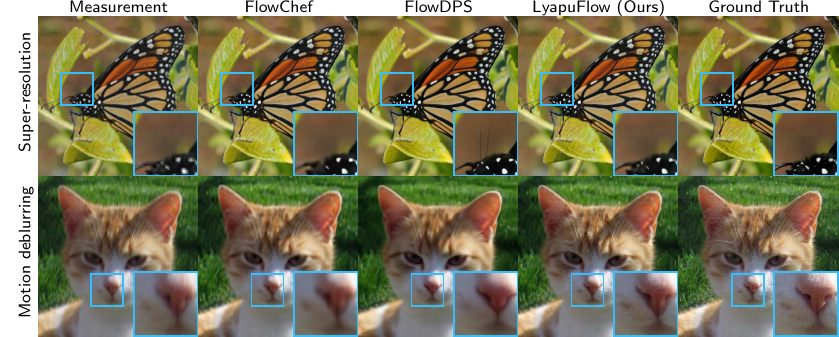}
  \caption{Qualitative comparison of reconstruction fidelity. LyapuFlow avoids hallucinated structures (top) while better preserving fine details (bottom).}
  \label{fig:main-compare-nfe28}
\end{figure}

\subsection{Image Inverse Problems}
\label{sec:experiment-image}

We next consider image inverse problems with measurement-consistency constraints. Given a degradation operator \(\mathcal A\) and noisy measurement \(\vy\), we define $\vh(\vx)=\mathcal A(\vx)-\vy$, so that satisfying \(\vh(\vx)=\mathbf 0\) corresponds to measurement consistency.
We test $\times4$ super-resolution, Gaussian deblurring, and motion deblurring across AFHQ, FFHQ, and DIV2K datasets using a pretrained latent flow model, Stable Diffusion 3.5-Medium~\citep{sd35}. This setting tests whether LyapuFlow can impose data consistency while preserving the instance-specific structure of the underlying image.

\Cref{tab:image-quantitative-nfe28} compares reconstruction quality across the three inverse problems under a matched $28$-NFE budget. LyapuFlow achieves the best or second-best performance across metrics, including the best SSIM and LPIPS throughout. This is achieved with a single feedback update per flow step, without iterative inner optimization. \Cref{fig:main-compare-nfe28} shows the same trend qualitatively. In super-resolution, FlowDPS introduces antenna-like structures on the butterfly that are absent from the target image, while in motion deblurring it produces a hazier reconstruction and removes distinctive markings around the cat's mouth. These examples suggest that measurement guidance can produce plausible reconstructions while losing instance-specific details. LyapuFlow better preserves these structures while maintaining measurement consistency. See Appendix~\ref{app:image-additional-results} for additional results.

\subsection{Ablation Studies} \label{sec:ablation}

\textbf{Role of LyapuFlow Components.} We isolate two components of the feedback construction: the explicit time derivative and the feedback trust region. \Cref{tab:lyapuflow-ablation-small} shows that removing the time-derivative term consistently degrades fidelity on Burgers and NS (with full tables in \cref{app:pde-ablation}), consistent with its role in tracking how the predicted terminal constraint changes along the flow. Empirically, removing the trust region generally degrades performance, consistent with its role in limiting large finite-step corrections. See \cref{app:image-additional-results} for ablations on the gradient switch step $k_\mathrm{sg}$.

\begin{table}[!t]
\centering
\caption{Ablations on feedback trust region and time derivative.}
\label{tab:lyapuflow-ablation-small}
\scriptsize
\setlength{\tabcolsep}{2pt}
\begin{tabular}{lrrrrrrrrrrr}
\toprule
 & \multicolumn{6}{c}{1D Burgers} & \multicolumn{5}{c}{2D NS} \\
\cmidrule(lr){2-7}\cmidrule(lr){8-12}
 Method& \(\frac{\mathrm{CMMSE}}{10^{-2}}\) & \(\frac{\mathrm{CSMSE}}{10^{-2}}\) & FPD & \(\frac{\mathrm{CE(IC)}}{10^{-2}}\) & \(\frac{\mathrm{CE(ND)}}{10^{-2}}\) & \(\frac{\mathrm{CE(CL)}}{10^{-2}}\) & \(\frac{\mathrm{CMMSE}}{10^{-2}}\) & \(\frac{\mathrm{CSMSE}}{10^{-2}}\) & FPD & \(\frac{\mathrm{CE(IC)}}{10^{-2}}\) & \(\frac{\mathrm{CE(CL)}}{10^{-2}}\) \\
\midrule
\rowcolor{LyapuCol} LyapuFlow (Ours) & \best{0.165} & \best{0.213} & \best{0.200} & 0 & \viooff{1.532} & \viooff{0.978} & \best{5.457} & \second{2.480} & \best{0.568} & 0 & 0 \\
w/o feedback trust region & \second{0.285} & \second{0.237} & \second{0.474} & 0 & \viooff{1.303} & \viooff{0.896} & \second{5.497} & \best{2.473} & \second{0.576} & 0 & 0 \\
w/o time derivative & 0.574 & 0.401 & 1.890 & 0 & \viooff{4.757} & \viooff{3.067} & 6.038 & 2.712 & 0.761 & 0 & 0 \\
\bottomrule
\end{tabular}
\end{table}

\begin{figure}[t!]
    \centering
    \includegraphics[width=0.92\linewidth]{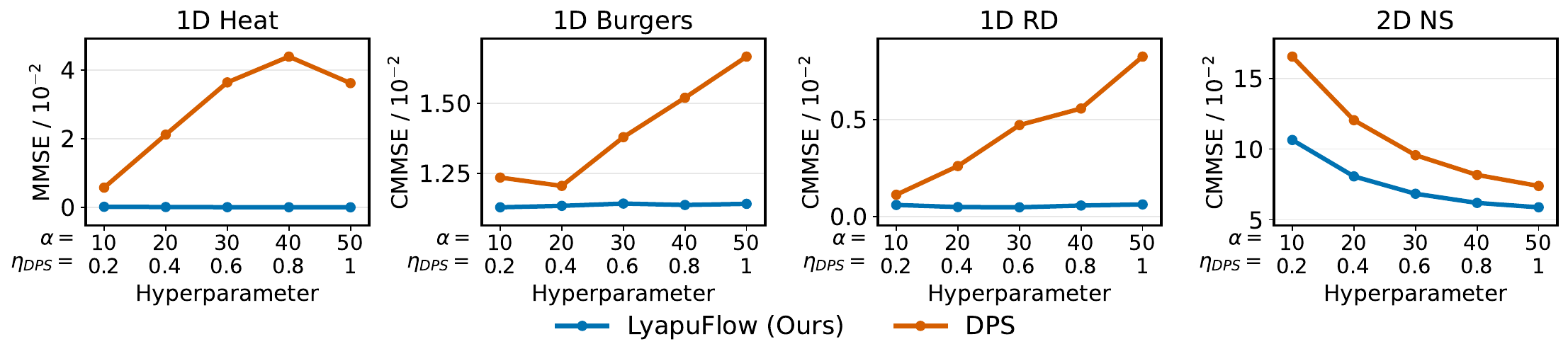}
    \vspace{-1.0em}
    \caption{Task adaptability of DPS and LyapuFlow to base step size and decay rate, respectively.}
    \label{fig:hparam_2samplers}
\end{figure}

\begin{figure}[!t]
  \centering
  \includegraphics[width=0.95\textwidth]{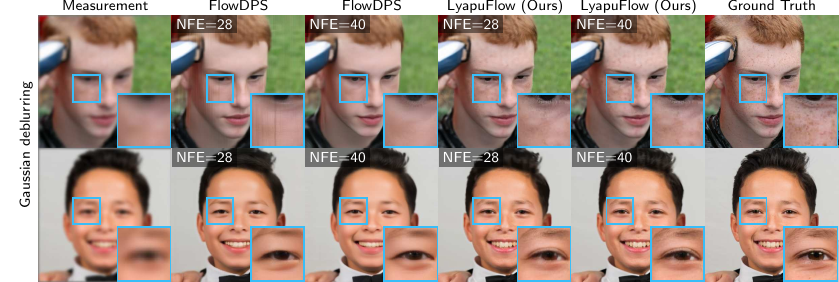}
  \caption{Qualitative comparison of reconstruction fidelity at different NFE budgets.}
  \label{fig:main-compare-nfe}
\end{figure}

\textbf{Hyperparameter Sensitivity.} We compare LyapuFlow and DPS through their principal guidance parameters. DPS directly tunes a step size $\eta_{\mathrm{DPS}}$, whereas LyapuFlow uses the decay rate $\alpha$ to determine the required feedback. \Cref{fig:hparam_2samplers} shows that the preferred DPS step size varies substantially across PDEs, with larger values helping NS but degrading Heat, Burgers, and RD. LyapuFlow is more stable across the tested values of $\alpha$, suggesting lower sensitivity to task-specific dynamics. Extended results for all samplers are provided in \cref{app:pde-sensitivity}.

\textbf{NFE Budget.} We first test whether the reconstruction gains persist as the integration budget increases. Figure~\ref{fig:main-compare-nfe} compares FlowDPS and LyapuFlow at $28$ and $40$ NFEs. Increasing the budget removes some FlowDPS artifacts, but does not recover instance-specific details lost during reconstruction. In the first row, facial freckles remain substantially weakened at both budgets, while in the second row the eyelid crease remains absent. LyapuFlow preserves these structures at both budgets. The additional evaluations at $40$ NFEs in~\cref{app:image-additional-results} show the same overall trend, indicating that the difference is not explained only by the number of integration steps.

\section{Conclusion}
We propose LyapuFlow, a test-time control framework for enforcing constraints on pretrained generative flows through Lyapunov feedback. By evaluating constraint violations on predicted terminal samples, LyapuFlow computes state-dependent corrections that are activated only when the pretrained dynamics fail to provide sufficient Lyapunov decrease. A feedback trust region further limits the control magnitude to reduce excessive finite-step corrections and the resulting discretization error. We demonstrate the approach in both data and latent spaces across scientific machine learning and image inverse problems. LyapuFlow consistently achieves strong constraint satisfaction and high sample fidelity relative to alternative test-time constraint-enforcement strategies.

\textbf{Limitation.} A current limitation is that LyapuFlow relies on differentiable constraint functions to construct the feedback signal. This restricts its direct applicability to constraints for which gradients with respect to the generated sample can be obtained. Extending LyapuFlow to non-differentiable, discrete, or black-box constraints is an interesting direction for future work. In addition, the proposed feedback introduces additional computational and memory overhead during generation. Combining LyapuFlow with acceleration techniques~\citep{SharpEuler, FastGen} is therefore another promising direction for future work.

\subsection*{Acknowledgments}
MG, HH, and NBE would like to acknowledge support from the U.S. Department of Energy, Office of Science, Office of Advanced Scientific Computing Research, through the EXPRESS: 2025 Exploratory Research for Extreme-Scale Science program and the RAPIDS Institute under the Scientific Discovery through Advanced Computing (SciDAC) program, under Contract No. DE-AC02-05CH11231 at Lawrence Berkeley National Laboratory. The views, opinions, and/or findings expressed are those of the authors and should not be interpreted as representing the official views or policies of the U.S. Government.

\bibliographystyle{plainnat}
\bibliography{references}

\clearpage
\appendix

\section*{Appendix Contents}

\startcontents[appendix]
\setcounter{tocdepth}{2}
\printcontents[appendix]{}{1}{}

\newpage

\section{Additional Analysis and Proofs}
\label{app:theory}

This appendix provides the derivations and additional conditions in \cref{sec:method}. We adapt standard tools from robust Lyapunov analysis and sampled-data nonlinear control to address the approximation introduced by LyapuFlow: the finite-step implementation of the feedback \cite{sontag1989universal,lin1996smooth,nevsic2005lyapunov,nevsic2006stabilization}. 

Throughout, all regularity and boundedness assumptions are required only on the region traversed by the trajectories under consideration. We assume that the pretrained velocity field $\vv_\theta(\vx,t)$ is sufficiently smooth for the corresponding flow map and derivatives below to exist, and that the constraint loss $\Ls$ is continuously differentiable. The discrete-time result requires the stronger smoothness condition stated separately in
\cref{app:discrete_stability}.

\subsection{Derivation of the Minimum-Norm Lyapunov Control}
\label{app:kkt-derivation}

We derive the minimum-norm control in \cref{eq:lyapuflow_control}, which is the optimal solution of
\begin{equation}
\begin{aligned}
\mathrm{minimize}&\;\;
\frac{1}{2}\|\vu_t\|_2^2
\\
\text{s.t.}&\;\;\nabla_{\vx_t}V^\top \vu_t
+
\partial_t V
+
\nabla_{\vx_t}V^\top \vv_\theta(\vx_t,t)
+
\alpha V
\leq 0.
\end{aligned}
\label{eq:app_min_norm_control}
\end{equation}
For brevity, define
\begin{equation}
\vg=\nabla_{\vx_t}V, \qquad b=\partial_t V +\nabla_{\vx_t}V^\top\vv_\theta(\vx_t,t) + \alpha V,
\end{equation}
so that the constraint becomes
\begin{equation}
\vg^\top\vu_t+b\leq0.
\end{equation}
For the corresponding Lagrangian
\begin{equation}
\mathcal J(\vu_t,\lambda)=\frac{1}{2}\|\vu_t\|_2^2 +\lambda(\vg^\top\vu_t+b),\qquad \lambda\geq0,
\end{equation}
the KKT conditions are
\begin{align}
\vu_t^\star+\lambda^\star\vg &= 0,
&& \text{(stationarity)}
\\
\vg^\top\vu_t^\star+b &\leq 0,
&& \text{(primal feasibility)}
\\
\lambda^\star &\geq 0,
&& \text{(dual feasibility)}
\\
\lambda^\star(\vg^\top\vu_t^\star+b) &= 0,
&& \text{(complementary slackness)}.
\end{align}
From stationarity,
\begin{equation} \label{eq:kkt-stationary}
\vu_t^\star=-\lambda^\star\vg.
\end{equation}
If $b\leq0$, an inactive control $\vu_t^\star=0$ is feasible and optimal. If $b>0$, assuming $\|\vg\|_2>0$ the constraint is active at the optimum, giving
\begin{equation}
-\lambda^\star\|\vg\|_2^2+b=0,
\end{equation}
and hence
\begin{equation}
\lambda^\star
=
\frac{b}{\|\vg\|_2^2}.
\end{equation}
Combining both cases,
\begin{equation}
\lambda^\star
=
\frac{[b]_+}{\|\vg\|_2^2}.
\end{equation}
By \cref{eq:kkt-stationary},
\begin{equation}
\vu_t^\star
=
-\frac{[b]_+}{\|\vg\|_2^2}\vg
=
-\frac{
\left[
\partial_t V
+
\nabla_{\vx_t}V^\top\vv_\theta(\vx_t,t)
+
\alpha V
\right]_+
}{
\|\nabla_{\vx_t}V\|_2^2
}
\nabla_{\vx_t}V,
\end{equation}
which proves \cref{eq:lyapuflow_control}.

\subsection{Discretization and Feedback Trust Region}
\label{app:discrete_stability}

We now consider the finite-step implementation of the continuous-time LyapuFlow sampler. Let
\begin{equation}
V_k = V(\vx_k,t_k),
\qquad
\vv_k = \vv_\theta(\vx_k,t_k),
\qquad
\Delta t_k = t_{k+1}-t_k > 0,
\end{equation}
and recall that the closed-form feedback can be written as
\begin{equation}
\vu_k^\star
=
-\eta_k\nabla_{\vx_k}V_k,
\end{equation}
where $\eta_k$ is the adaptive feedback gain defined in
\cref{eq:lyapuflow_control}. Let $\tilde{\vu}_k$ denote the feedback actually
applied by the numerical sampler. The Euler update is
\begin{equation}
\vx_{k+1}
=
\vx_k
+
\Delta t_k
\left(
\vv_k+\tilde{\vu}_k
\right).
\label{eq:appendix_euler_update}
\end{equation}
The feedback therefore contributes the finite displacement
$\Delta t_k\tilde{\vu}_k$. Before any feedback bound is applied, its magnitude
is
\begin{equation}
\left\|
\Delta t_k\vu_k^\star
\right\|_2
=
\Delta t_k\eta_k
\left\|
\nabla_{\vx_k}V_k
\right\|_2.
\label{eq:appendix_feedback_displacement}
\end{equation}
Thus, even though $\eta_k$ is chosen to satisfy an instantaneous
continuous-time condition, the actual correction made by the numerical
sampler also depends on the integration step and the local gradient magnitude.

As in the main text, we measure any shortfall in the requested
continuous-time Lyapunov decrease by
\begin{equation}
\xi_k
=
\left[
\partial_tV_k
+
\nabla_{\vx_k}V_k^\top
\left(
\vv_k+\tilde{\vu}_k
\right)
+
\alpha V_k
\right]_+.
\label{eq:appendix_clf_slack}
\end{equation}
Equivalently,
\begin{equation}
\partial_tV_k
+
\nabla_{\vx_k}V_k^\top
\left(
\vv_k+\tilde{\vu}_k
\right)
\leq
-\alpha V_k+\xi_k.
\label{eq:appendix_slack_inequality}
\end{equation}
If $\tilde{\vu}_k=\vu_k^\star$, the closed-form controller satisfies the
Lyapunov condition and $\xi_k=0$. Bounding the feedback can change the applied
control and introduce a positive shortfall.

\paragraph{Proof of \cref{prop:discrete_stability}.}
Define the augmented state
\begin{equation}
\vz_k
=
\begin{bmatrix}
\vx_k\\
t_k
\end{bmatrix},
\qquad
\Delta\vz_k
=
\Delta t_k
\begin{bmatrix}
\vv_k+\tilde{\vu}_k\\
1
\end{bmatrix}.
\end{equation}
Assume that the gradient of $V$ with respect to the augmented variable $(\vx,t)$ is $L_V$-Lipschitz on a neighborhood containing the line segment between $\vz_k$ and $\vz_{k+1}$. The descent lemma gives
\begin{align}
V_{k+1}-V_k
\leq
\Delta t_k
\left[
\partial_tV_k
+
\nabla_{\vx_k}V_k^\top
\left(
\vv_k+\tilde{\vu}_k
\right)
\right]
+
\frac{L_V}{2}
\|\Delta\vz_k\|_2^2.
\label{eq:appendix_descent_lemma}
\end{align}
Since
\begin{equation}
\|\Delta\vz_k\|_2^2
=
\Delta t_k^2
\left(
1+\|\vv_k+\tilde{\vu}_k\|_2^2
\right),
\end{equation}
substituting \cref{eq:appendix_slack_inequality} yields
\begin{equation}
V_{k+1}-V_k
\leq
-\alpha\Delta t_k V_k
+
\Delta t_k\xi_k
+
\frac{L_V}{2}\Delta t_k^2
\left(
1+\|\vv_k+\tilde{\vu}_k\|_2^2
\right),
\end{equation}
which proves \cref{prop:discrete_stability}.

The bound separates two effects of modifying the feedback. Restricting the
control can increase $\xi_k$ and therefore weaken the instantaneous Lyapunov
decrease. Allowing a large control, on the other hand, increases the
second-order term through the magnitude of the controlled velocity. This is
the finite-step trade-off underlying the feedback trust region.

\paragraph{One-step contraction.}
The same bound gives a sufficient condition for preserving contraction after
an Euler step. For any $\gamma\in(0,1)$, suppose
\begin{equation}
\xi_k
+
\frac{L_V}{2}\Delta t_k
\left(
1+\|\vv_k+\tilde{\vu}_k\|_2^2
\right)
\leq
(1-\gamma)\alpha V_k.
\label{eq:appendix_discrete_condition}
\end{equation}
Then
\begin{equation}
V_{k+1}
\leq
\left(
1-\gamma\alpha\Delta t_k
\right)V_k.
\label{eq:appendix_discrete_contraction}
\end{equation}
For $0<\gamma\alpha\Delta t_k<1$ and $V_k>0$, the Euler step therefore
retains a strict contraction.

This condition also illustrates why aggressive feedback can be problematic
near a nearly satisfied constraint. Even when $\xi_k=0$, the finite-step
error increases with
$\Delta t_k\|\vv_k+\tilde{\vu}_k\|_2$, whereas the right-hand side of
\cref{eq:appendix_discrete_condition} decreases with $V_k$. A feedback law
that is valid instantaneously can therefore fail to reduce $V$ after a coarse
finite step.

\paragraph{Accumulated discretization error.}
Consider a uniform grid with $\Delta t=1/K$. Suppose
\begin{equation}
\xi_k\leq\bar{\xi},
\qquad
\|\vv_k\|_2\leq B_v,
\qquad
\|\tilde{\vu}_k\|_2\leq B_u
\end{equation}
for all $k$, and assume $0<\alpha\Delta t<1$. Then
\cref{prop:discrete_stability} gives
\begin{equation}
V_{k+1}
\leq
(1-\alpha\Delta t)V_k
+
\Delta t\,\bar{\xi}
+
\frac{L_V}{2}\Delta t^2
\left(
1+(B_v+B_u)^2
\right).
\label{eq:appendix_uniform_recurrence}
\end{equation}
Iterating the recurrence gives
\begin{align}
V_K
\leq\;&
(1-\alpha\Delta t)^K V_0
\nonumber\\
&+
\left[
\frac{\bar{\xi}}{\alpha}
+
\frac{L_V\Delta t}{2\alpha}
\left(
1+(B_v+B_u)^2
\right)
\right]
\left[
1-(1-\alpha\Delta t)^K
\right].
\label{eq:appendix_terminal_discrete_bound}
\end{align}
When $\bar{\xi}=0$, the additional error introduced by Euler integration is
$\mathcal{O}(\Delta t)=\mathcal{O}(1/K)$. Its constant grows with the
magnitude of the applied feedback. Increasing the number of integration steps
and increasing the allowed feedback magnitude therefore have coupled effects
on the discrete sampler.

If $\bar{\xi}>0$, grid refinement still removes the Euler discretization term,
but the contribution $\bar{\xi}/\alpha$ need not vanish. This distinction is
important: reducing $\Delta t$ controls numerical integration error, whereas
persistent relaxation of the Lyapunov condition introduces a separate source
of error.

\paragraph{Effect of the feedback trust region.}
The feedback trust region in \cref{eq:control-trust-region} bounds the
finite displacement caused specifically by the feedback,
\begin{equation}
\left\|
\Delta t_k\tilde{\vu}_k
\right\|_2
\leq
\rho_k.
\end{equation}
We choose
\begin{equation}
\rho_k
=
\Delta t_k \bar{u}_k,
\end{equation}
which is equivalent to
$\|\tilde{\vu}_k\|_2\leq \bar{u}_k$. With fixed $\bar{u}_k$, the feedback
displacement therefore vanishes as the integration grid is refined. In
contrast, keeping $\rho_k$ fixed independently of $\Delta t_k$ would permit
$\|\tilde{\vu}_k\|_2=\mathcal{O}(1/\Delta t_k)$, so the finite-step error in
\cref{prop:discrete_stability} would not necessarily vanish under grid
refinement.

To make the effect of clipping explicit, write
\begin{equation}
\tilde{\vu}_k
=
\beta_k\vu_k^\star,
\qquad
\tilde{\eta}_k
=
\beta_k\eta_k,
\label{eq:appendix_clipped_gain}
\end{equation}
where
\begin{equation}
\beta_k
=
\begin{cases}
1,
&
\|\vu_k^\star\|_2\leq \bar{u}_k,
\\[3pt]
\dfrac{\bar{u}_k}{\|\vu_k^\star\|_2},
&
\|\vu_k^\star\|_2>\bar{u}_k.
\end{cases}
\label{eq:appendix_clipping_factor}
\end{equation}
Hence
\begin{equation}
\tilde{\vu}_k
=
-\tilde{\eta}_k\nabla_{\vx_k}V_k,
\end{equation}
and the trust region acts by reducing the adaptive feedback gain while
preserving its direction.

The corresponding Lyapunov shortfall can be computed exactly. Let
$\vg_k=\nabla_{\vx_k}V_k$. From the definition of $\eta_k$,
\begin{equation}
\eta_k\|\vg_k\|_2^2
=
\left[
\partial_tV_k
+
\vg_k^\top\vv_k
+
\alpha V_k
\right]_+.
\label{eq:appendix_gain_identity}
\end{equation}
Substituting
$\tilde{\vu}_k=-\beta_k\eta_k\vg_k$
into \cref{eq:appendix_clf_slack} gives
\begin{equation}
\xi_k
=
(1-\beta_k)\eta_k\|\vg_k\|_2^2
=
(\eta_k-\tilde{\eta}_k)\|\vg_k\|_2^2.
\label{eq:appendix_clipping_slack}
\end{equation}
Thus, when the trust region is inactive, $\beta_k=1$ and the requested
instantaneous decrease is preserved. When clipping is active, the trust region
reduces the finite feedback displacement
\begin{equation}
\Delta t_k
\|\tilde{\vu}_k\|_2
=
\Delta t_k
\tilde{\eta}_k
\|\vg_k\|_2
\end{equation}
at the cost of the explicitly characterized shortfall in
\cref{eq:appendix_clipping_slack}. This makes the role of the feedback trust
region precise: it moderates the effective finite-step correction rather than
changing the direction selected by the Lyapunov controller.

\paragraph{Degenerate gradients.}
The closed-form controller assumes that the gradient of $V$ provides a
direction through which the control can change the Lyapunov derivative. If
$\nabla_{\vx}V=\vzero$ while
\begin{equation}
\partial_tV+\alpha V>0,
\end{equation}
then the control does not affect $\dot V$ to first order and the requested
Lyapunov decrease cannot be achieved by any finite instantaneous control.
Our analysis therefore assumes
$\nabla_{\vx}V\neq\vzero$ whenever a nonzero correction is required. If
$\nabla_{\vx}V=\vzero$ and the Lyapunov condition is already satisfied, we
take $\vu^\star=\vzero$.

\newpage
\section{Physics-Constrained Generation}
\label{app:pde}

This section provides the complete experimental setup and additional results for physics-constrained generation. We first describe the PDE tasks, pretrained generative backbones, test-time constraints, and evaluation protocol. We then provide additional analyses of computational cost, hyperparameter sensitivity, and the components of LyapuFlow, followed by expanded qualitative results.

\subsection{Experimental Setup}
\label{app:pde-setup}

\subsubsection{Tasks and Datasets}
\label{app:pde-tasks-datasets}

Following ECI and PCFM~\citep{ECI,PCFM}, we consider four PDE benchmarks spanning different dynamics and constraint structures: 1D Heat, 1D Burgers, 1D Reaction--Diffusion (RD), and 2D Navier--Stokes (NS). We use $s$ for spatial coordinates and $\tau$ for physical time, and $\mathcal{U}$ denotes the uniform distribution. \Cref{tab:data} summarizes the quantitative data information.

\begin{table}[!ht]
\centering
\caption{Summary of PDE datasets.}
\label{tab:data}
\footnotesize
\begin{tabular}{lrrrr}
\toprule
Task & Spatial dimension & Temporal dimension & $d$ & Train samples\\
\midrule
1D Heat & $100$ & $100$ & $10{,}000$ & 5,000 \\
1D Burgers & $101$ & $101$ & $10{,}201$ & 6,400 \\
1D RD & $128$ & $100$ & $12{,}800$ & 6,400 \\
2D NS & $64\times64$ & $50$ & $204{,}800$ & 10,000 \\
\bottomrule
\end{tabular}
\end{table}

\paragraph{1D Heat.}
On the spatial domain $s\in[0,2\pi)$ and time interval $\tau\in[0,1]$, the dataset is generated analytically as
\begin{equation}
x(s,\tau)=\sin(s+\phi)\exp(-\nu\tau),
\qquad
\phi\sim\mathcal{U}(0,\pi),
\quad
\nu\sim\mathcal{U}(1,5).
\end{equation}
This is an exact solution of $x_{\tau}=\nu x_{ss}$ with periodic boundary conditions. The spatial and temporal grids contain 100 points each, and each trajectory is generated from a newly sampled $(\phi,\nu)$ pair.

\paragraph{1D Burgers.}
The inviscid Burgers equation on $s,\tau\in[0,1]$ is solved with a Godunov flux,
\begin{equation}
x_{\tau}+\partial_s\left(\tfrac12 x^2\right)=0,
\qquad
x(s,0)=\frac{1}{1+\exp((s-p)/0.02)},
\qquad
p\sim\mathcal{U}(0.2,0.8).
\end{equation}
One hundred spatial and temporal intervals yield a $101\times101$ stored field. The parameter $p$ determines the initial field used as the test-time initial-condition target.

\paragraph{1D Reaction--Diffusion.}
The dataset is generated from
\begin{equation}
x_{\tau}=\nu x_{ss}+\rho x(1-x),
\qquad
\rho=0.01,
\quad
\nu=0.005.
\end{equation}
The spatial domain contains 128 cell-centered points in $[0,1]$, with 100 stored times from 0 to 0.99. The left and right diffusive fluxes are sampled as $g_L\in[0,0.05]$ and $g_R\in[-0.05,0]$. Initial fields are normalized random combinations of up to three Fourier modes, with occasional sign, absolute-value, and compact-support transformations.

\paragraph{2D Navier--Stokes.}
On the unit two-torus, let $\mathbf{s}=(s_1,s_2)$ denote the spatial coordinate. The vorticity equation is
\begin{equation}
\partial_\tau\omega
+
\mathbf v\cdot\nabla_{\mathbf s}\omega
=
\mu\Delta_{\mathbf s}\omega+f,
\qquad
\Delta_{\mathbf s}\psi=\omega,
\qquad
\mathbf v=(\partial_{s_2}\psi,-\partial_{s_1}\psi),
\qquad
\mu=10^{-3}.
\end{equation}
Initial vorticity is sampled from a two-dimensional Gaussian random field. The forcing family is
$f_j(\mathbf s)=0.1\sqrt2\sin(2\pi(s_1+s_2)+\phi_j)$
for 100 phases $\phi_j\in[0,\pi/2]$. Fourier-space Crank--Nicolson integration with $2/3$ dealiasing and internal step $10^{-3}$ runs to time 49 and records 50 snapshots.

\subsubsection{Generative Backbone}
\label{app:pde-backbone}

We use unconditional functional flow matching (FFM) backbones~\citep{ffm} so that the physical conditions imposed at test time are not available to the generative model during training. Each PDE-specific model uses an Fourier neural operator (FNO)~\citep{fno} to learn a velocity field $v_\theta(x,t)$ that transports a functional prior $x_0\sim p_0$ to a solution $x_1\sim p_{\rm data}$. For $t\sim\mathcal{U}(0,1)$, training uses
\begin{equation}
x_t=(1-t)x_0+t x_1,
\qquad
v_t=x_1-x_0,
\qquad
\mathcal L_{\rm FFM}
=
\mathbb E\left[\lVert v_\theta(x_t,t)-v_t\rVert^2\right].
\end{equation}
Initial conditions, boundary conditions, and forcings are not provided as explicit conditioning channels. The backbone instead models the distribution of complete solution fields induced by their variation.

The Heat, Burgers, and RD models use four Fourier layers, while the NS model uses two. For Heat, Burgers, and RD, $p_0$ is a Mat\'ern Gaussian-process prior with length scale $0.001$ and variance $1$; NS uses an i.i.d.\ Gaussian prior. Heat, Burgers, and RD are trained with Adam using learning rate $3\times10^{-4}$, batch size $256$, and $20$k optimization steps. NS is trained with AdamW using the same learning rate, batch size $24$, and $500$k optimization steps.

\subsubsection{Test-Time Constraints}
\label{app:pde-constraints}

At inference time, each task introduces physical information that was not provided to the unconditional backbone during training. The constraints include direct conditions on the solution, such as an initial field, together with conservation, balance, or local-dynamics residuals that couple multiple parts of the generated trajectory. This provides a range of constraint dimensions and nonlinearities across the four PDEs.

\Cref{tab:constraints} summarizes the residuals. Each constraint maps a flattened solution field $x\in\R^d$ to a residual $h(x)\in\R^m$, with quadratic constraint loss
$\Ls(x)=\tfrac12\lVert h(x)\rVert^2$.
For Burgers, the local nonlinear-dynamics residual applies the Godunov update over the first five time transitions. Its mass-balance residual accounts for transport through the boundaries using the Burgers flux $F(x)=x^2/2$ and trapezoidal integration in time, rather than requiring total mass to remain constant. 
For RD, we define the mass \(M^n=\int_\Omega x(s,\tau_n)\,ds\)., the spatially integrated reaction source \(S^n=\int_\Omega \rho x(s,\tau_n)(1-x(s,\tau_n))\,ds\), and the boundary diffusion contribution \(\Phi^n=\nu[x_s(s_R,\tau_n)-x_s(s_L,\tau_n)]\).
For NS, the zero-mean forcing motivates a total-vorticity conservation constraint.

\begin{table}[!ht]
    \centering
    \footnotesize
    \caption{Summary of the test-time PDE constraints.}
    \label{tab:constraints}
    \resizebox{\textwidth}{!}{
    \begin{tabular}{llrl}
        \toprule
        Task & Type & $m$ & Residual\\
        \midrule
        \multirow{3}{*}{1D Heat}
        & Linear IC & 100 & $x(:,0)-x^{\rm ref}(:,0)$\\
        & Linear CL & 99 & $\sum_s x(s,\tau)\Delta s-\sum_s x(s,0)\Delta s$, $\tau=1,\ldots,99$\\
        & Total & 199 & \\
        \midrule
        \multirow{4}{*}{1D Burgers}
        & Linear IC & 101 & $x(:,0)-x^{\rm ref}(:,0)$\\
        & Nonlinear Dynamics & 495 & $\mathcal{R}_j^n=x_j^{n+1}-x_j^n+\frac{\Delta\tau}{\Delta s}[G(x_j^n,x_{j+1}^n)-G(x_{j-1}^n,x_j^n)]$, $n=0,\ldots,4$\\
        & Nonlinear CL & 100 & $M^n-M^0+\Delta\tau\sum_{q=0}^{n-1}(\Phi^q+\Phi^{q+1})/2$, $n=1,\ldots,100$\\
        & Total & 696 & \\
        \midrule
        \multirow{3}{*}{1D RD}
        & Linear IC & 128 & $x(:,0)-x^{\rm ref}(:,0)$\\
        & Nonlinear CL & 99 & \(M^n-M^0 - \Delta\tau\sum_{q=0}^{n-1}\frac{(\Phi^q+S^q)+(\Phi^{q+1}+S^{q+1})}{2},\qquad n=1,\ldots,100.\) \\
        & Total & 227 & \\
        \midrule
        \multirow{3}{*}{2D NS}
        & Linear IC & $4,096$ & $\omega(:,:,0)-\omega_0$\\
        & Linear CL & 49 & $\sum_{s_1,s_2}\omega(\tau)\Delta s_1\Delta s_2-\sum_{s_1,s_2}\omega(0)\Delta s_1\Delta s_2$, $\tau=1,\ldots,49$\\
        & Total & 4,145 & \\
        \bottomrule
    \end{tabular}}
\end{table}

\subsection{Baselines and Evaluation Protocol}
\label{app:pde-protocol}

We compare LyapuFlow with samplers that intervene at different points along the generative trajectory. DPS~\citep{DPS} and SAD-Flower~\citep{SADFlower} correct intermediate states $x_t$; FlowDPS~\citep{FlowDPS}, ProFlow~\citep{ProFlow}, and PCFM~\citep{PCFM} operate on terminal predictions $\hat{x}_{1|t}$; and OC-Flow~\citep{OCFlow} optimizes the complete trajectory $x_{[0,1]}$. DPS denotes its flow-based variant and provides a comparison that, like LyapuFlow, differentiates through the pretrained velocity model.

For a controlled comparison, all samplers use the same final projection after generation. The FFM reference performs unconditional sampling followed by this projection, without an intermediate test-time correction. We tune the principal sampler-specific hyperparameters on each task and report the selected values below. The evaluation configurations and method-specific hyperparameters are summarized in the following tables.

\begin{table}[!ht]
\centering
\caption{Construction of the test-reference data used for PDE simulation. For each selected condition, reference trajectories are fetched from the indicated complementary condition set.}
\label{tab:pde-test-data}
\scriptsize
\setlength{\tabcolsep}{3pt}
\renewcommand{\arraystretch}{1.08}
\begin{tabular}{llll}
\toprule
\multicolumn{1}{c}{Dataset} & \multicolumn{1}{c}{Selected conditions} & \multicolumn{1}{c}{Reference fetch per condition} & \multicolumn{1}{c}{Total} \\
\midrule
1D Heat & 1 IC (\(\phi=\pi/4\)) & 512 analytic \(\nu\)-trajectories, \(\nu\sim\mathcal{U}[1,5]\) & 512 \\
\midrule
1D Reaction--Diffusion & 16 IC & 32 BC trajectories per IC & 512 \\
\midrule
1D Burgers & 16 IC & 32 BC trajectories per IC & 512 \\
\midrule
2D Navier--Stokes & 10 IC (vorticity) & 10 forcing trajectories per IC (vorticity) & 100 \\
\bottomrule
\end{tabular}
\end{table}

\begin{table}[!ht]
\centering
\caption{Hyperparameters used in physics-constrained generation. Hyperparameter sweeps use \(N=64\) samples per dataset and \(N_{\mathrm{idx}}\in\{16,10\}\) selected indices for multi-condition datasets. \(\alpha\) and \(\alpha_{\mathrm{SF}}\) are the LyapuFlow and SAD-Flower gains, respectively; \(\eta_{\mathrm{DPS}}\) is the DPS base step size for $\eta_\mathrm{DPS}/\|\vh(\vx)\|$; \(K_{iter}\) is the number of iterations; \(\eta_{\mathrm{FD}}\), \(\eta_{\mathrm{PF}}\), and \(\eta_{\mathrm{OC}}\) are the FlowDPS, ProFlow, and OC-Flow step sizes, respectively.}
\label{tab:pde-hparams}
\scriptsize
\setlength{\tabcolsep}{3pt}
\renewcommand{\arraystretch}{1.08}
\resizebox{\textwidth}{!}{%
\begin{tabular}{lllllll}
\toprule
Dataset & \multicolumn{1}{c}{LyapuFlow} & \multicolumn{1}{c}{DPS} & \multicolumn{1}{c}{SAD-Flower} & \multicolumn{1}{c}{FlowDPS} & \multicolumn{1}{c}{ProFlow} & \multicolumn{1}{c}{OC-Flow} \\
\midrule
1D Heat & \(\alpha=50\); \(c_{\mathrm{tr}}=10\) & \(\eta_{\mathrm{DPS}}=0.2\) & \(\alpha_{\mathrm{SF}}=14\) & \(K_{iter}=3\); \(\eta_{\mathrm{FD}}=0.02\) & \(K_{iter}=3\); \(\eta_{\mathrm{PF}}=0.1\) & \(K_{iter}=3\); \(\beta=0.995\); \(\eta_{\mathrm{OC}}=0.5\) \\
\midrule
1D Burgers & \(\alpha=10\); \(c_{\mathrm{tr}}=1\) & \(\eta_{\mathrm{DPS}}=0.4\) & \(\alpha_{\mathrm{SF}}=5\) & \(K_{iter}=3\); \(\eta_{\mathrm{FD}}=0.06\) & \(K_{iter}=3\); \(\eta_{\mathrm{PF}}=0.1\) & \(K_{iter}=3\); \(\beta=0.995\); \(\eta_{\mathrm{OC}}=2.5\) \\
\midrule
1D RD & \(\alpha=30\); \(c_{\mathrm{tr}}=1\) & \(\eta_{\mathrm{DPS}}=0.2\) & \(\alpha_{\mathrm{SF}}=5\) & \(K_{iter}=3\); \(\eta_{\mathrm{FD}}=0.06\) & \(K_{iter}=3\); \(\eta_{\mathrm{PF}}=0.5\) & \(K_{iter}=3\); \(\beta=0.995\); \(\eta_{\mathrm{OC}}=0.5\) \\
\midrule
2D NS & \(\alpha=50\); \(c_{\mathrm{tr}}=1\) & \(\eta_{\mathrm{DPS}}=1\) & \(\alpha_{\mathrm{SF}}=14\) & \(K_{iter}=3\); \(\eta_{\mathrm{FD}}=0.08\) & \(K_{iter}=3\); \(\eta_{\mathrm{PF}}=0.5\) & \(K_{iter}=3\); \(\beta=0.995\); \(\eta_{\mathrm{OC}}=0.5\) \\
\bottomrule
\end{tabular}
}
\end{table}

\subsection{Evaluation Metrics}
\label{app:pde-metrics}

A partial physical condition generally leaves other initial, boundary, or forcing variables unspecified and therefore does not determine a unique PDE trajectory. Each condition instead defines a distribution of admissible solutions. We evaluate whether the generated samples recover this conditional distribution in addition to satisfying the imposed constraints.

For a single condition group, MMSE compares the pointwise sample-mean fields,
\begin{equation}
\mathrm{MMSE} = \frac{1}{d}
\lVert \bar{x}-\bar{x}^{\rm ref} \rVert^2, 
\qquad
\mathrm{SMSE} = \frac{1}{d}
\lVert \sigma_x-\sigma_x^{\rm ref} \rVert^2.
\end{equation}
where
$\bar{x}=\frac{1}{N}\sum_{i=1}^{N}x_i$ and
$\bar{x}^{\rm ref}=\frac{1}{N}\sum_{i=1}^{N}x_i^{\rm ref}$.
SMSE compares the corresponding pointwise standard-deviation fields.

For Burgers, RD, and NS, we evaluate multiple partial conditions and report macro-averaged conditional metrics. Given $G$ condition groups with index sets $I_g$,
\begin{equation}
\mathrm{CMMSE}
=
\frac{1}{G}
\sum_{g=1}^{G}
\frac{1}{d}
\lVert
\bar{x}_g-\bar{x}^{\rm ref}_g
\rVert^2, \qquad
\mathrm{CSMSE}
=
\frac{1}{G}
\sum_{g=1}^{G}
\frac{1}{d}
\lVert
\sigma_{x,g}-\sigma^{\rm ref}_{x,g}
\rVert^2.
\end{equation}
Here, $\bar{x}_g$ and $\sigma_{x,g}$ denote the group-wise mean and standard-deviation fields, with analogous definitions for the reference samples.

Constraint error measures the mean residual norm of individual generated samples,
\begin{equation}
\mathrm{CE}
=
\frac{1}{N}
\sum_i
\lVert h(x_i)\rVert.
\end{equation}

We additionally report Fr\'echet Poseidon Distance (FPD) as a feature-space measure of distributional similarity. Generated and reference fields are embedded using the PDE foundation model Poseidon--B~\citep{poseidon}, and the Fr\'echet distance is computed from the resulting Gaussian feature statistics. For the 1D datasets, each spatiotemporal field is processed as one feature. For NS, each time frame is embedded separately and the frame-wise distances are averaged. The $128\times100$ RD fields are center-cropped to $100\times100$ before feature extraction to match the model input shape.

Together, CE measures physical feasibility, while CMMSE, CSMSE, and FPD measure complementary aspects of fidelity to the conditional solution distribution.

\subsection{Sampling Efficiency}
\label{app:pde-efficiency}

The evaluated samplers differ substantially in how much computation is required to impose a test-time constraint. \Cref{tab:sampling-efficiency} compares per-sample runtime and peak GPU memory, including the common final projection used in the evaluation. FlowDPS, ProFlow, and OC-Flow use \(K=3\) inner iterations in the reported configuration, so additional iterations would further increase their cost. OC-Flow is the most expensive method across all four tasks, consistent with its repeated optimization of the complete control trajectory. PCFM also incurs additional cost from Jacobian evaluation and linear-system solves. LyapuFlow avoids trajectory-level iterative optimization and remains computationally competitive while applying constraint-aware feedback throughout sampling.

\begin{table}[!ht]
\centering
\caption{Sampling efficiency measured on the same NVIDIA A100 GPU with batch size 1 and 100 sampling steps. Wall time is the mean elapsed time per generated sample over 5 timed runs after 1 warm-up run; peak allocated memory is the maximum CUDA allocation across those runs. Both measurements include one-step final projection.}
\label{tab:sampling-efficiency}
\scriptsize
\setlength{\tabcolsep}{3pt}
\renewcommand{\arraystretch}{1.08}
\resizebox{\textwidth}{!}{%
\begin{tabular}{llrrrrrrr}
\toprule
Dataset & Metric \(\downarrow\) & LyapuFlow (Ours) & DPS & SAD-Flower & FlowDPS & ProFlow & PCFM & OC-Flow \\
\midrule
\multirow{2}{*}{1D Heat} & Second / sample & \cellcolor{efficiencyred!11}1.106 & \cellcolor{efficiencyred!6}0.802 & \cellcolor{efficiencyred!0}0.401 & \cellcolor{efficiencyred!5}0.71 & \cellcolor{efficiencyred!6}0.774 & \cellcolor{efficiencyred!3}0.618 & \cellcolor{efficiencyred!40}2.992 \\
 & Peak memory (GiB) & \cellcolor{efficiencyred!3}0.167 & \cellcolor{efficiencyred!2}0.157 & \cellcolor{efficiencyred!0}0.115 & \cellcolor{efficiencyred!0}0.115 & \cellcolor{efficiencyred!40}0.833 & \cellcolor{efficiencyred!21}0.493 & \cellcolor{efficiencyred!24}0.545 \\
\midrule
\multirow{2}{*}{1D Burgers} & Second / sample & \cellcolor{efficiencyred!21}1.87 & \cellcolor{efficiencyred!12}1.29 & \cellcolor{efficiencyred!0}0.579 & \cellcolor{efficiencyred!27}2.242 & \cellcolor{efficiencyred!28}2.295 & \cellcolor{efficiencyred!19}1.712 & \cellcolor{efficiencyred!40}3 \\
 & Peak memory (GiB) & \cellcolor{efficiencyred!2}0.168 & \cellcolor{efficiencyred!1}0.158 & \cellcolor{efficiencyred!0}0.141 & \cellcolor{efficiencyred!0}0.141 & \cellcolor{efficiencyred!40}0.862 & \cellcolor{efficiencyred!27}0.62 & \cellcolor{efficiencyred!23}0.561 \\
\midrule
\multirow{2}{*}{1D RD} & Second / sample & \cellcolor{efficiencyred!15}1.351 & \cellcolor{efficiencyred!8}0.959 & \cellcolor{efficiencyred!0}0.475 & \cellcolor{efficiencyred!14}1.297 & \cellcolor{efficiencyred!15}1.375 & \cellcolor{efficiencyred!9}0.987 & \cellcolor{efficiencyred!40}2.867 \\
 & Peak memory (GiB) & \cellcolor{efficiencyred!2}0.186 & \cellcolor{efficiencyred!2}0.174 & \cellcolor{efficiencyred!0}0.119 & \cellcolor{efficiencyred!0}0.119 & \cellcolor{efficiencyred!40}1.31 & \cellcolor{efficiencyred!21}0.748 & \cellcolor{efficiencyred!23}0.805 \\
\midrule
\multirow{2}{*}{2D NS} & Second / sample & \cellcolor{efficiencyred!12}1.985 & \cellcolor{efficiencyred!6}1.505 & \cellcolor{efficiencyred!0}1.008 & \cellcolor{efficiencyred!3}1.281 & \cellcolor{efficiencyred!4}1.325 & \cellcolor{efficiencyred!30}3.468 & \cellcolor{efficiencyred!40}4.277 \\
 & Peak memory (GiB) & \cellcolor{efficiencyred!0}6.382 & \cellcolor{efficiencyred!0}6.382 & \cellcolor{efficiencyred!0}6.382 & \cellcolor{efficiencyred!0}6.382 & \cellcolor{efficiencyred!0}6.382 & \cellcolor{efficiencyred!40}9.607 & \cellcolor{efficiencyred!0}6.382 \\
\bottomrule
\end{tabular}
}
\end{table}

\newpage
\subsection{Component Ablations}
\label{app:pde-ablation}

We isolate the two components examined in the main-text ablation, the explicit time derivative in the Lyapunov dynamics and the feedback trust region. \Cref{tab:lyapuflow-ablation} shows that removing the time-derivative term generally increases fidelity error, with a particularly clear effect on FPD. This is consistent with its role in accounting for changes in the predicted terminal constraint along the pretrained flow. Removing the trust region has a smaller effect, with performance remaining close to the full method on several metrics. This supports its role as a safeguard against large finite-step corrections rather than as the primary source of the feedback signal.

\begin{table}[!ht]
\centering
\caption{Ablations on feedback trust region and time derivative across all PDE tasks.}
\label{tab:lyapuflow-ablation}
\scriptsize
\setlength{\tabcolsep}{3pt}
\renewcommand{\arraystretch}{0.95} 
\resizebox{\textwidth}{!}{%
\begin{tabular}{lrrrrrrrrrrr}
\toprule
 & \multicolumn{5}{c}{1D Heat} & \multicolumn{6}{c}{1D Burgers} \\
\cmidrule(lr){2-6}\cmidrule(lr){7-12}
Method & \(\frac{\mathrm{MMSE}}{10^{-2}}\) & \(\frac{\mathrm{SMSE}}{10^{-2}}\) & FPD & \(\frac{\mathrm{CE(IC)}}{10^{-2}}\) & \(\frac{\mathrm{CE(CL)}}{10^{-2}}\) & \(\frac{\mathrm{CMMSE}}{10^{-2}}\) & \(\frac{\mathrm{CSMSE}}{10^{-2}}\) & FPD & \(\frac{\mathrm{CE(IC)}}{10^{-2}}\) & \(\frac{\mathrm{CE(ND)}}{10^{-2}}\) & \(\frac{\mathrm{CE(CL)}}{10^{-2}}\) \\
\midrule
\rowcolor{LyapuCol} LyapuFlow (Ours) & \best{0.012} & \best{0.004} & \best{0.006} & 0 & 0 & \best{0.163} & \best{0.212} & \best{0.204} & 0 & 0.010 & 0.005 \\
w/o trust region & \best{0.012} & \best{0.004} & \best{0.006} & 0 & 0 & \second{0.283} & \second{0.235} & \second{0.458} & 0 & 0.008 & 0.004 \\
w/o time derivative & \second{0.067} & \second{0.018} & \second{0.043} & 0 & 0 & 0.584 & 0.398 & 2.240 & 0 & 2.208 & 4.800 \\
\bottomrule
\end{tabular}
}
\resizebox{\textwidth}{!}{%
\begin{tabular}{lrrrrrrrrrr}
\toprule
 & \multicolumn{5}{c}{1D RD} & \multicolumn{5}{c}{2D NS} \\
\cmidrule(lr){2-6}\cmidrule(lr){7-11}
Method & \(\frac{\mathrm{CMMSE}}{10^{-2}}\) & \(\frac{\mathrm{CSMSE}}{10^{-2}}\) & FPD & \(\frac{\mathrm{CE(IC)}}{10^{-2}}\) & \(\frac{\mathrm{CE(NCL)}}{10^{-2}}\) & \(\frac{\mathrm{CMMSE}}{10^{-2}}\) & \(\frac{\mathrm{CSMSE}}{10^{-2}}\) & FPD & \(\frac{\mathrm{CE(IC)}}{10^{-2}}\) & \(\frac{\mathrm{CE(CL)}}{10^{-2}}\) \\
\midrule
\rowcolor{LyapuCol} LyapuFlow (Ours) & \best{0.019} & \best{0.028} & \best{0.002} & 0 & 0 & \best{5.457} & \second{2.480} & \best{0.568} & 0 & 0 \\
w/o trust region & \best{0.019} & \best{0.028} & \best{0.002} & 0 & 0 & \second{5.497} & \best{2.473} & \second{0.576} & 0 & 0 \\
w/o time derivative & \second{0.052} & \second{0.029} & \second{3.525} & 0 & 0 & 6.038 & 2.712 & 0.761 & 0 & 0 \\
\bottomrule
\end{tabular}
}
\end{table}

\subsection{Hyperparameter Sensitivity}
\label{app:pde-sensitivity}

\begin{figure}[!ht]
    \centering
    \includegraphics[width=\linewidth]{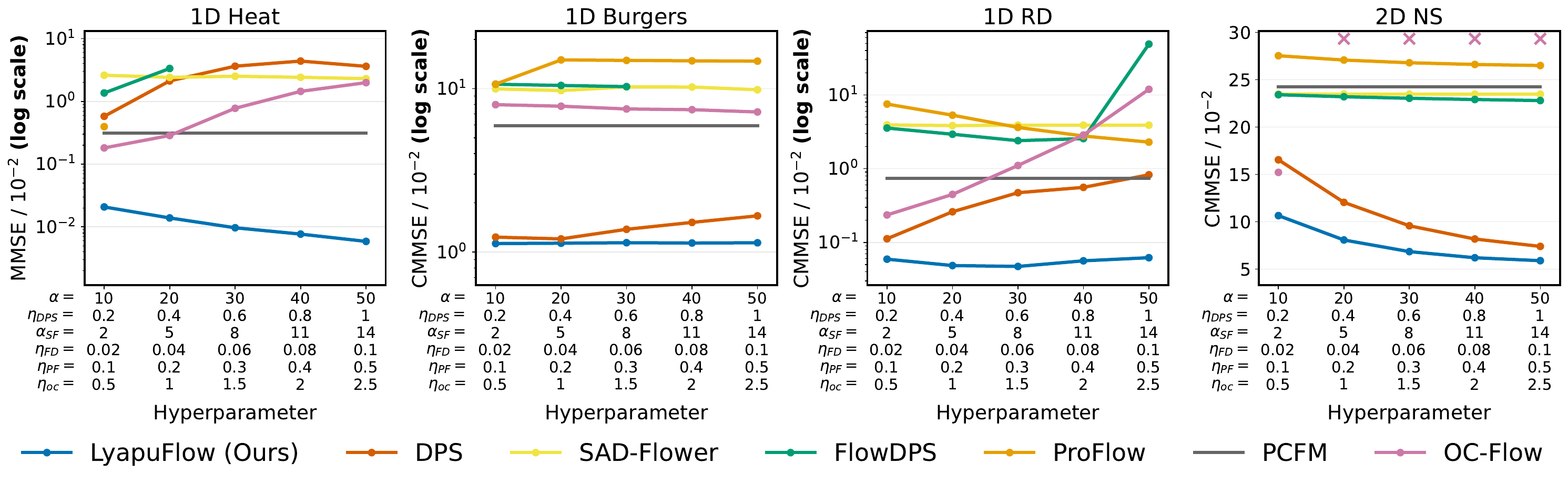}
    \vspace{-2.5em}
    \caption{Hyperparameter sensitivity across the evaluated PDE samplers. Each method is evaluated over five values of its principal guidance parameter: $\alpha$ for LyapuFlow, $\eta_{\mathrm{DPS}}$ for DPS, $\alpha_{\mathrm{SF}}$ for SAD-Flower, $\eta_{\mathrm{FD}}$ for FlowDPS, $\eta_{\mathrm{PF}}$ for ProFlow, and $\eta_{\mathrm{OC}}$ for OC-Flow. PCFM has no tuned guidance parameter and is shown as a horizontal reference. The 1D tasks use a logarithmic vertical scale, and non-finite evaluations are marked with $\times$.}
    \label{fig:hparam_7samplers}
\end{figure}

We evaluate whether the principal guidance parameter of each sampler transfers reliably across PDE dynamics or requires substantial task-specific adjustment. \Cref{fig:hparam_7samplers} varies one parameter at a time while keeping the remaining settings fixed. PCFM has no corresponding tunable guidance parameter and is shown as a horizontal reference.

For each method, we first identify a coarse range that avoids non-finite evaluations and then tune within that range. Methods that directly parameterize the correction magnitude can still become unstable for some candidate values. LyapuFlow instead tunes the decay rate $\alpha$, which specifies the requested Lyapunov decrease while the feedback magnitude is determined from the current state. Across the evaluated tasks, LyapuFlow varies less across the tested values than methods whose principal parameter directly controls a correction step size. The comparison with DPS in \cref{fig:hparam_2samplers} provides the corresponding linear-scale view used in the main paper.

\clearpage
\subsection{Qualitative Results}
\label{app:pde-vis}

The following examples complement the aggregate metrics by showing how the different samplers modify complete PDE solution fields under the imposed test-time conditions.

\begin{figure}[!ht]
    \centering
    \includegraphics[page=1,width=0.9\textwidth]{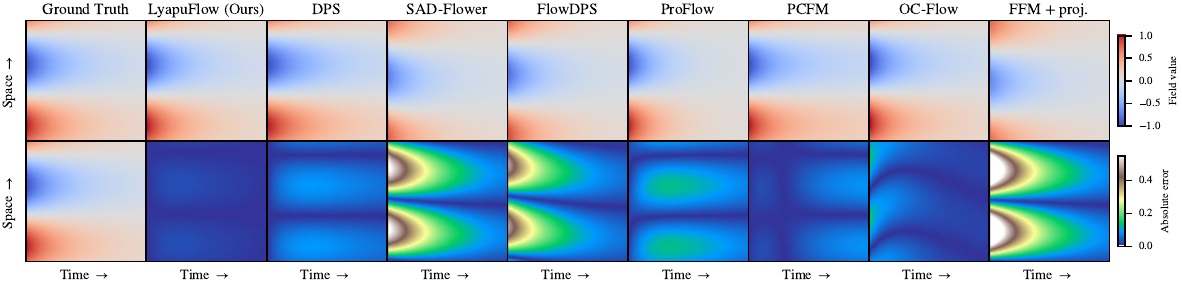}
    \caption{Qualitative comparison on 1D Heat under initial-condition and conservation-law constraints.}
    \label{fig:pde_heat}
\end{figure}

\begin{figure}[!ht]
    \centering
    \includegraphics[page=1,width=0.9\textwidth]{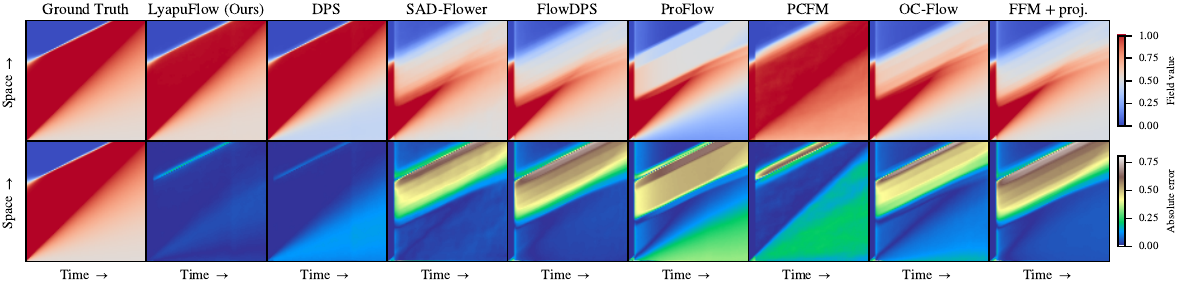}
    \includegraphics[page=8,width=0.9\textwidth]{figs/PDE/burgers1d_ic_average.pdf}
    \includegraphics[page=14,width=0.9\textwidth]{figs/PDE/burgers1d_ic_average.pdf}
    \includegraphics[page=9,width=0.9\textwidth]{figs/PDE/burgers1d_ic_average.pdf}
    \caption{Qualitative comparisons on 1D Burgers under initial-condition, local-dynamics, and mass-balance constraints.}
    \label{fig:pde_burgers1d_ic}
\end{figure}

\begin{figure}[!ht]
    \centering
    \includegraphics[page=2,width=\textwidth]{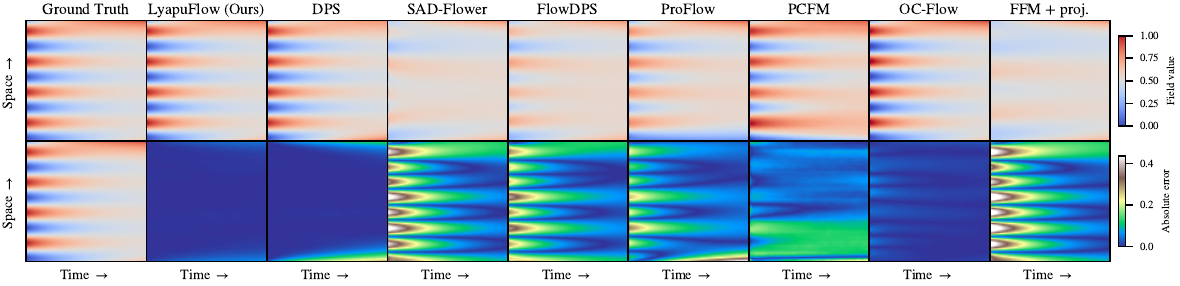}
    \includegraphics[page=3,width=\textwidth]{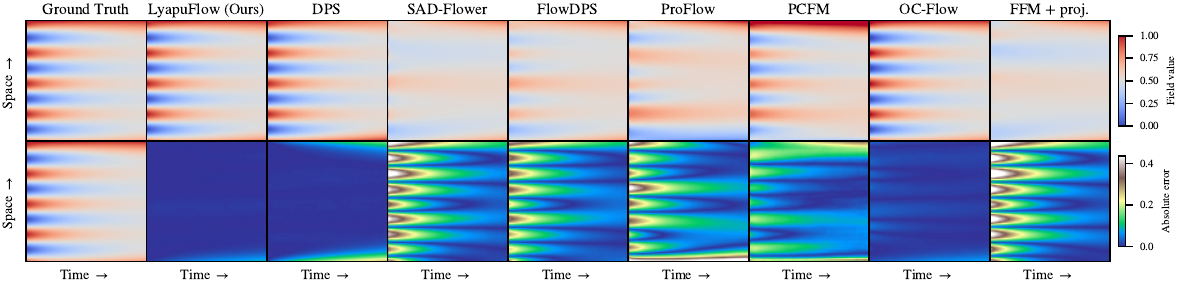}
    \includegraphics[page=5,width=\textwidth]{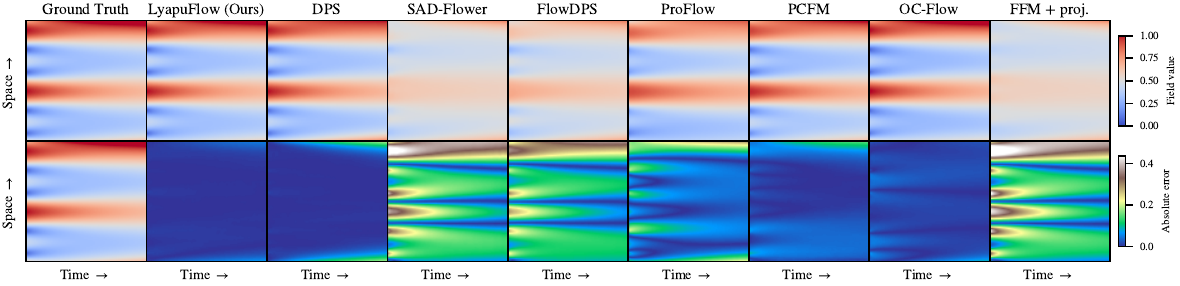}
    \includegraphics[page=12,width=\textwidth]{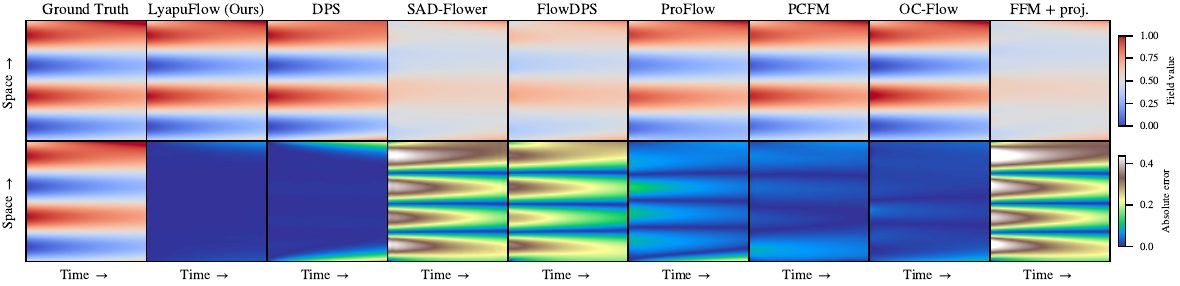}
    \caption{Qualitative comparisons on 1D Reaction--Diffusion under initial-condition and balance-law constraints.}
    \label{fig:pde_rd1d}
\end{figure}

\begin{figure}[!ht]
    \centering
    \includegraphics[page=6,width=\textwidth]{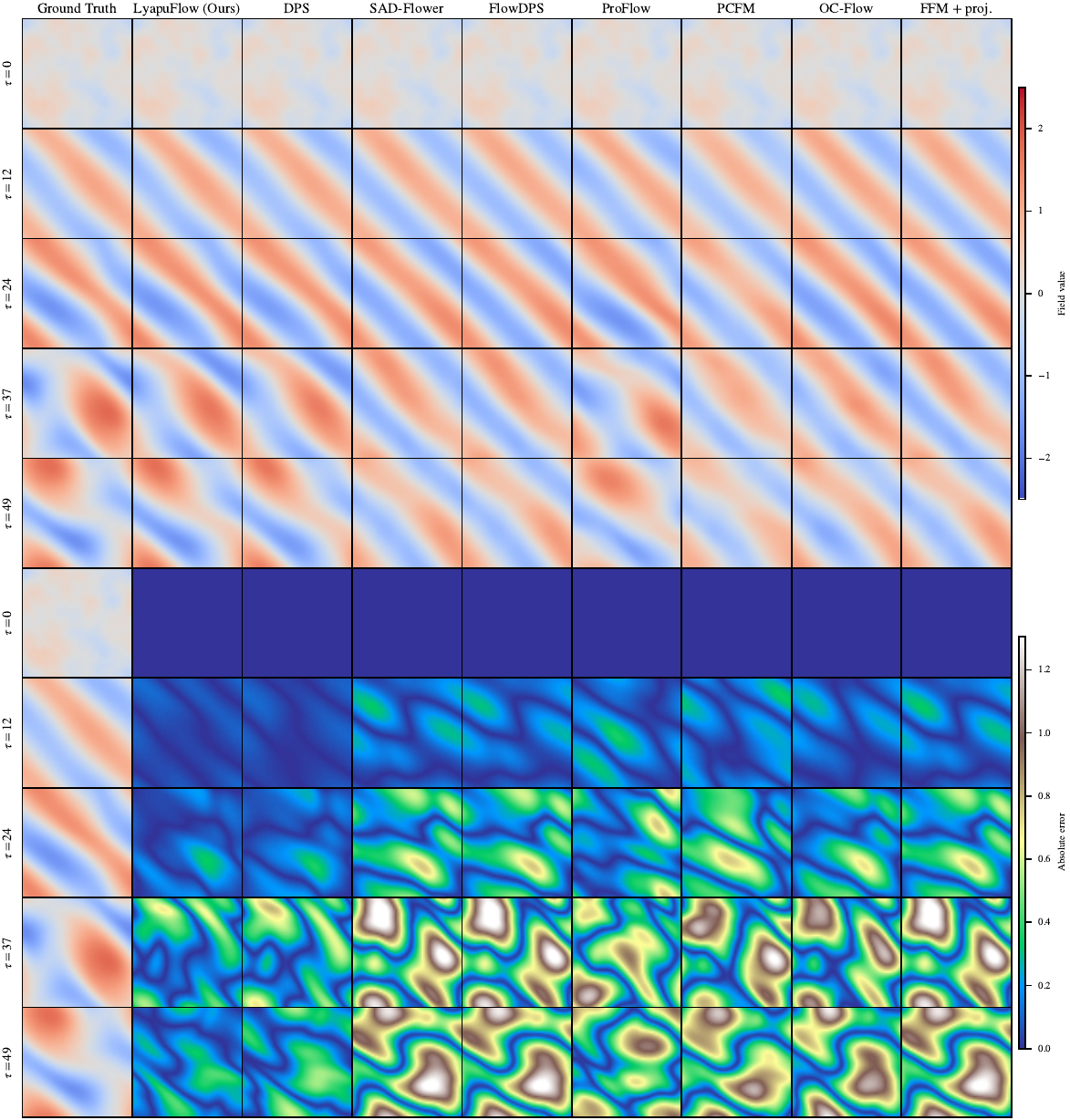}
    \caption{Qualitative comparisons on 2D Navier--Stokes under initial-vorticity and total-vorticity constraints.}
    \label{fig:pde_ns_1}
\end{figure}
\begin{figure}[!ht]
    \centering
    \includegraphics[page=7,width=\textwidth]{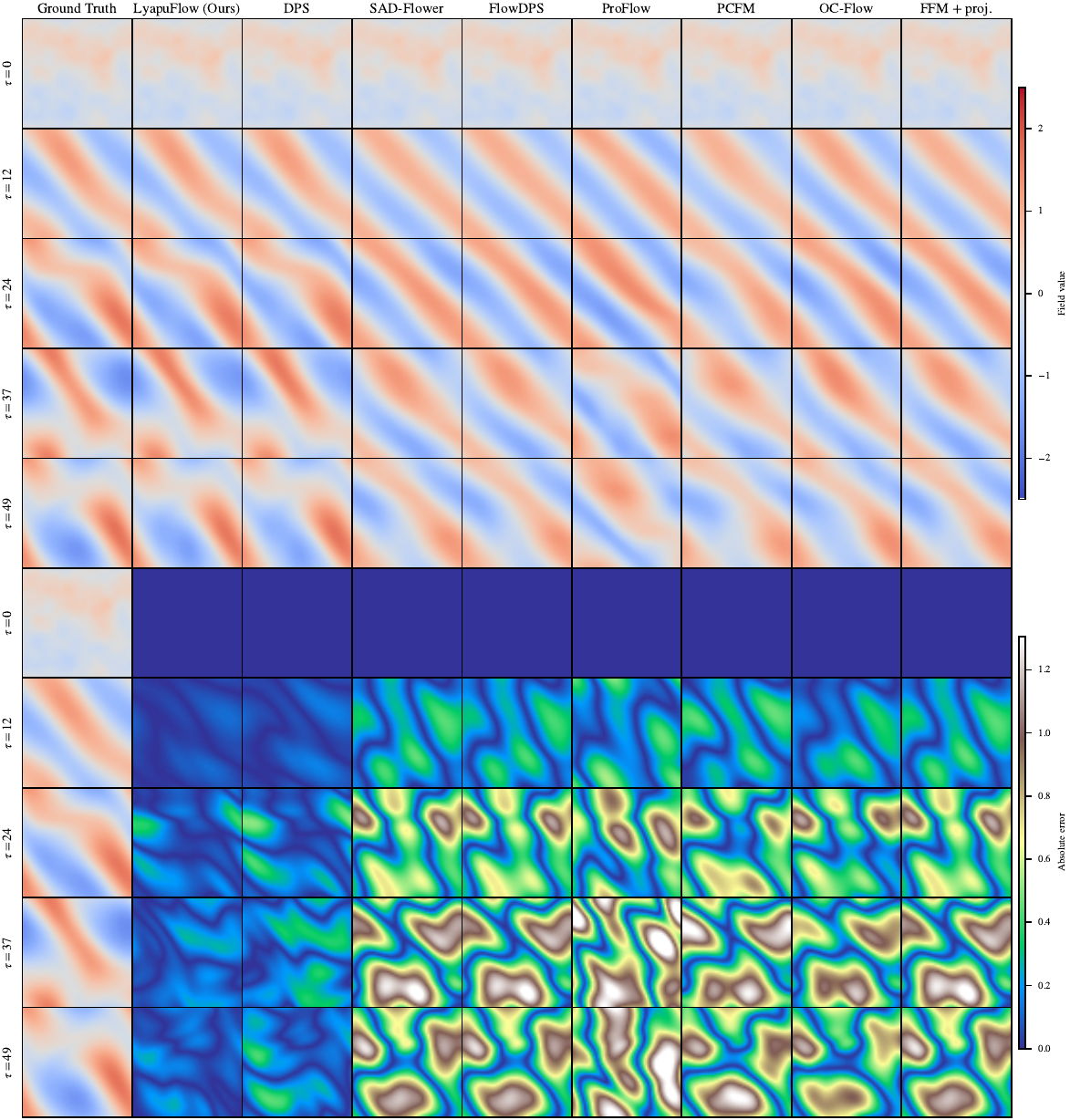}

    \caption{Qualitative comparisons on 2D Navier--Stokes under initial-vorticity and total-vorticity constraints.}
    \label{fig:pde_ns_2}
\end{figure}

\clearpage
\section{Image Inverse Problems}
\label{app:image}

This section provides the complete experimental setup and additional results for latent-space image inverse problems. We first describe the datasets, pretrained backbone, degradation operators, baseline configurations, and evaluation protocol. We then examine sampling efficiency, performance at a larger integration budget, frequency-domain behavior, and additional qualitative reconstructions.

\subsection{Experimental Setup}
\label{app:image-setup}

\subsubsection{Datasets and Backbone}
\label{app:image-data-backbone}

We evaluate on $1$k AFHQ images, comprising $0.5$k dog and $0.5$k cat images, $1$k FFHQ images, and $0.8$k DIV2K images, all at $768\times768$ resolution. All methods use Stable Diffusion 3.5-Medium~\citep{sd35} as the pretrained latent-space flow model with classifier-free guidance scale $2$~\citep{cfg}. AFHQ uses the prompts ``a photo of a dog'' and ``a photo of a cat'', FFHQ uses ``a photo of a person,'' and DIV2K uses ``a high quality photo of \{DAPE tags\}'', with the tags extracted from the DAPE-degraded images~\citep{DAPE}.

\subsubsection{Inverse Problems and Observations}
\label{app:image-tasks}

We consider $\times4$ bicubic super-resolution, Gaussian deblurring, and motion deblurring with kernel size $61$. For each task, the observed image is obtained by applying the corresponding forward operator and adding measurement noise with $\sigma_n=0.03$. The resulting observation defines the test-time data-consistency constraint. The same pretrained latent flow is therefore reused across all inverse problems, while only the observation operator and measurement change at inference time.

\subsection{Baselines and Evaluation Protocol}
\label{app:image-protocol}

We consider LatentDAPS~\citep{DAPS}, FlowChef~\citep{FlowChef}, FlowDPS~\citep{FlowDPS}, FLAIR~\citep{FLAIR}, and FlowLPS~\citep{FlowLPS} as candidate baselines. The primary main-text comparison focuses on FlowChef and FlowDPS because, like LyapuFlow, they operate without the much larger inner optimization loops used by several of the other methods. We report the computational cost of all candidate baselines separately in \cref{app:image-efficiency}.

\paragraph{Method-Specific Settings.}
\begin{itemize}[leftmargin=*]
    \item \textbf{LatentDAPS~\citep{DAPS}.}
    $20$ EDM annealing steps from $\sigma=1$ to $0.2$ with $\rho=7$; each interval uses two-step midpoint denoising, for $40$ total NFEs. Non-SR tasks use $15$ Langevin MCMC updates per annealing step with learning rate $10^{-4}$. Super-resolution uses $60$ updates with learning rate $10^{-3}$, decayed by $0.7$ every $5$ MCMC updates; $\tau=0.03$.

    \item \textbf{FlowChef~\citep{FlowChef}.}
    One data-consistency update per flow step with step size $15$.

    \item \textbf{FlowDPS~\citep{FlowDPS}.}
    Three data-consistency updates per flow step with step size $15$.

    \item \textbf{FLAIR~\citep{FLAIR}.}
    Official repository defaults with $15$ data-term updates per step, data-term SGD learning rate $1.0$ for super-resolution and $0.1$ for deblurring, regularizer scale $0.5$, the bundled calibration schedule, and $t_{\min}=0.18$.

    \item \textbf{FlowLPS~\citep{FlowLPS}.}
    $N_L=5$ Langevin and $N_P=10$ proximal-SGD updates. The Langevin step size is $10^{-4}$, the proximal-SGD learning rate is $0.1$ without decay, the schedule offset is $3$, and pCN re-noising is used.

    \item \textbf{LyapuFlow.}
    See \cref{tab:image-hparams}.
\end{itemize}

\begin{table}[!ht]
    \centering
    \small
    \caption{Task-specific LyapuFlow settings for 28 and 40 NFEs.
    $k_{\mathrm{sg}}$ denotes the control step at which stop-gradient on
    $v_\theta$ is turned off (zero-indexed); $k_{\mathrm{sg}}$ equal to the
    number of NFEs keeps it on throughout sampling.}
    \label{tab:image-hparams}
    \begin{tabular}{llrrr@{\hspace{0.6cm}}rrr}
    \toprule
    & &
    \multicolumn{3}{c}{{28 NFEs}}
    &
    \multicolumn{3}{c}{{40 NFEs}}
    \\
    \cmidrule(lr){3-5}
    \cmidrule(lr){6-8}
    Dataset & Task
    & $\alpha$ & $k_{\mathrm{sg}}$ & $c_{\mathrm{tr}}$
    & $\alpha$ & $k_{\mathrm{sg}}$ & $c_{\mathrm{tr}}$
    \\
    \midrule
    AFHQ & Super-resolution (Bicubic)
    & 5 & 28 & 1.0
    & 5 & 40 & 1.0
    \\
    AFHQ & Deblurring (Gaussian)
    & 5 & 3 & 1.0
    & 5 & 40 & 1.0
    \\
    AFHQ & Deblurring (Motion)
    & 5 & 7 & 1.0
    & 5 & 40 & 1.0
    \\
    \midrule
    FFHQ & Super-resolution (Bicubic)
    & 5 & 28 & 1.0
    & 5 & 40 & 1.0
    \\
    FFHQ & Deblurring (Gaussian)
    & 5 & 3 & 1.0
    & 5 & 7 & 1.0
    \\
    FFHQ & Deblurring (Motion)
    & 5 & 7 & 1.0
    & 5 & 7 & 1.0
    \\
    \midrule
    DIV2K & Super-resolution (Bicubic)
    & 5 & 28 & 1.0
    & 5 & 40 & 1.0
    \\
    DIV2K & Deblurring (Gaussian)
    & 5 & 3 & 1.0
    & 5 & 40 & 1.0
    \\
    DIV2K & Deblurring (Motion)
    & 7 & 7 & 1.0
    & 5 & 40 & 1.0
    \\
    \bottomrule
    \end{tabular}
\end{table}

\subsection{Evaluation Metrics}
\label{app:image-metrics}

We report peak signal-to-noise ratio (PSNR) and structural similarity index (SSIM) to measure reconstruction distortion and structural agreement. We additionally report Learned Perceptual Image Patch Similarity (LPIPS) and Fr\'echet Inception Distance (FID) to capture perceptual and distributional differences that are not fully reflected by pixel-space distortion.

\subsection{Sampling Efficiency}
\label{app:image-efficiency}

The candidate image baselines differ substantially in the amount of computation performed within each sampling step. \Cref{tab:image-first10speed} compares reconstruction quality and runtime on a small evaluation subset. The iteration-heavy methods LatentDAPS, FLAIR, and FlowLPS incur substantially greater sampling cost than single-update methods. For example, FlowLPS improves PSNR over LyapuFlow by $1.21$ dB on motion deblurring while requiring more than $8\times$ longer sampling time. This motivates the primary main-text comparison with FlowChef and FlowDPS, which operates in a more comparable quality--efficiency regime.

\begin{table}[!ht]
    \small
    \centering
    \caption{Trade-off between sampling speed and quality, evaluated on the first 10 FFHQ images.}    
    \label{tab:image-first10speed}
    \setlength{\tabcolsep}{2pt}
    \begin{tabular}{l c c *{6}{r} c}
    \toprule
    Sampler & $N_{iter}$ & NFE & \multicolumn{2}{c}{SR $\times 4$} & \multicolumn{2}{c}{Deblur (Gaussian)} & \multicolumn{2}{c}{Deblur (Motion)} & Second / image \\
    \cmidrule(lr){4-5} \cmidrule(lr){6-7} \cmidrule(lr){8-9}
    & & & PSNR & LPIPS & PSNR & LPIPS & PSNR & LPIPS & \\
    \midrule
    LyapuFlow (Ours) & 1 & 40 & \underline{30.63} & \textbf{0.041} & \underline{28.76} & 0.117 & \underline{28.87} & 0.120 &  \underline{13.8} \\
    FlowChef & 1 & 40 & \textbf{30.90} & 0.060 & 26.62 & 0.219 & 28.52 & 0.175 &  \textbf{11.9} \\
    FlowDPS & 3 & 40 & 30.09 & 0.101 & 26.57 & 0.208 & 27.83 & 0.171 &  26.6 \\
    \midrule
    FlowLPS & 15 & 40 & 24.86 & 0.060 & \textbf{28.87} & \textbf{0.075} & \textbf{30.08} & \textbf{0.065} & \vio 112.6 \\
    FLAIR & 15 & 40 & 22.84 & \underline{0.044} & 26.29 & \underline{0.094} & 26.93 & \underline{0.088} & \vio 112.8 \\
    LatentDAPS & 15 or 60 & 40 & 27.05 & 0.116 & 25.22 & 0.238 & 25.29 & 0.249 & \vio 112.6 \\
    \bottomrule
    \end{tabular}
\end{table}

\clearpage
\subsection{Additional Results and Ablations}
\label{app:image-additional-results}

We additionally evaluate the methods at $40$ NFEs to test whether the differences observed at $28$ NFEs are explained by a limited integration budget. The results follow the same overall trend as the main evaluation, with LyapuFlow maintaining strong reconstruction quality as the number of flow evaluations increases.

\begin{table*}[!ht]
    \centering
    \caption{Quantitative evaluation for image inverse problems. All experiments use Stable Diffusion 3.5-Medium with 40 NFEs. Bold: best; Underline: second best.}    
    \label{tab:image-quantitative-nfe40}

    \resizebox{\linewidth}{!}{
    \begin{tabular}{lcccccccccccc}
    \toprule
    \rowcolor{Gray}
    & \multicolumn{12}{c}{AFHQ 1k (768 $\times$ 768)}\\
    \multirow{2}{*}{Method} & \multicolumn{4}{c}{Super-resolution x4 (Bicubic)} & \multicolumn{4}{c}{Deblurring (Gauss)} & \multicolumn{4}{c}{Deblurring (Motion)}\\
    \cmidrule(lr){2-5} \cmidrule(lr){6-9} \cmidrule(lr){10-13}
    & PSNR$\uparrow$ & SSIM$\uparrow$ & FID$\downarrow$ & LPIPS$\downarrow$ & PSNR$\uparrow$ & SSIM$\uparrow$ & FID$\downarrow$ & LPIPS$\downarrow$ & PSNR$\uparrow$ & SSIM$\uparrow$ & FID$\downarrow$ & LPIPS$\downarrow$ \\
    \midrule
    FlowChef & \underline{29.67} & \underline{0.807} & 22.34 & \underline{0.077} & 25.42 & \underline{0.719} & 38.55 & 0.256 & \underline{26.56} & \underline{0.743} & 36.37 & 0.2107 \\
    FlowDPS & 28.51 & 0.764 & \underline{18.45} & 0.124 & \underline{25.67} & 0.697 & \underline{29.65} & \underline{0.244} & 26.20 & 0.710 & \underline{25.21} & \underline{0.2106} \\
    LyapuFlow (Ours) & \textbf{30.35} & \textbf{0.833} & \textbf{8.93} & \textbf{0.040} & \textbf{27.42} & \textbf{0.732} & \textbf{17.86} & \textbf{0.137} & \textbf{27.57} & \textbf{0.750} & \textbf{17.06} & \textbf{0.1429} \\
    \midrule
    \rowcolor{Gray}
    & \multicolumn{12}{c}{FFHQ 1k (768 $\times$ 768)}\\
    \multirow{2}{*}{Method} & \multicolumn{4}{c}{Super-resolution x4 (Bicubic)} & \multicolumn{4}{c}{Deblurring (Gauss)} & \multicolumn{4}{c}{Deblurring (Motion)}\\
    \cmidrule(lr){2-5} \cmidrule(lr){6-9} \cmidrule(lr){10-13}
    & PSNR$\uparrow$ & SSIM$\uparrow$ & FID$\downarrow$ & LPIPS$\downarrow$ & PSNR$\uparrow$ & SSIM$\uparrow$ & FID$\downarrow$ & LPIPS$\downarrow$ & PSNR$\uparrow$ & SSIM$\uparrow$ & FID$\downarrow$ & LPIPS$\downarrow$ \\
    \midrule
    FlowChef & \underline{29.99} & \underline{0.835} & 61.02 & \underline{0.061} & 25.78 & \underline{0.763} & 113.95 & 0.219 & \underline{27.13} & \textbf{0.786} & 94.87 & 0.179 \\
    FlowDPS & 29.39 & 0.778 & \underline{28.45} & 0.098 & \underline{26.29} & 0.739 & \underline{50.23} & \underline{0.206} & 27.09 & 0.747 & \underline{40.28} & \underline{0.173} \\
    LyapuFlow (Ours) & \textbf{30.28} & \textbf{0.836} & \textbf{26.48} & \textbf{0.044} & \textbf{28.04} & \textbf{0.770} & \textbf{40.20} & \textbf{0.122} & \textbf{28.21} & \underline{0.782} & \textbf{35.86} & \textbf{0.123} \\
    \midrule
    \rowcolor{Gray}
    & \multicolumn{12}{c}{DIV2K 0.8k (768 $\times$ 768)}\\
    \multirow{2}{*}{Method} & \multicolumn{4}{c}{Super-resolution x4 (Bicubic)} & \multicolumn{4}{c}{Deblurring (Gauss)} & \multicolumn{4}{c}{Deblurring (Motion)}\\
    \cmidrule(lr){2-5} \cmidrule(lr){6-9} \cmidrule(lr){10-13}
    & PSNR$\uparrow$ & SSIM$\uparrow$ & FID$\downarrow$ & LPIPS$\downarrow$ & PSNR$\uparrow$ & SSIM$\uparrow$ & FID$\downarrow$ & LPIPS$\downarrow$ & PSNR$\uparrow$ & SSIM$\uparrow$ & FID$\downarrow$ & LPIPS$\downarrow$ \\
    \midrule
    FlowChef & \underline{24.48} & \underline{0.683} & 43.25 & \underline{0.087} & \underline{20.84} & \underline{0.536} & 95.28 & \underline{0.327} & \underline{21.94} & \underline{0.581} & 80.43 & \underline{0.263} \\
    FlowDPS & 24.07 & 0.637 & \underline{32.96} & 0.113 & 20.81 & 0.505 & \underline{89.69} & 0.334 & 21.66 & 0.541 & \underline{73.33} & 0.269 \\
    LyapuFlow (Ours) & \textbf{24.82} & \textbf{0.710} & \textbf{23.35} & \textbf{0.047} & \textbf{22.24} & \textbf{0.564} & \textbf{68.25} & \textbf{0.211} & \textbf{22.74} & \textbf{0.607} & \textbf{61.19} & \textbf{0.204} \\
    \bottomrule
    \end{tabular}
    }
\end{table*}

\begin{figure}[!ht]
    \centering
    \includegraphics[width=0.8\linewidth]{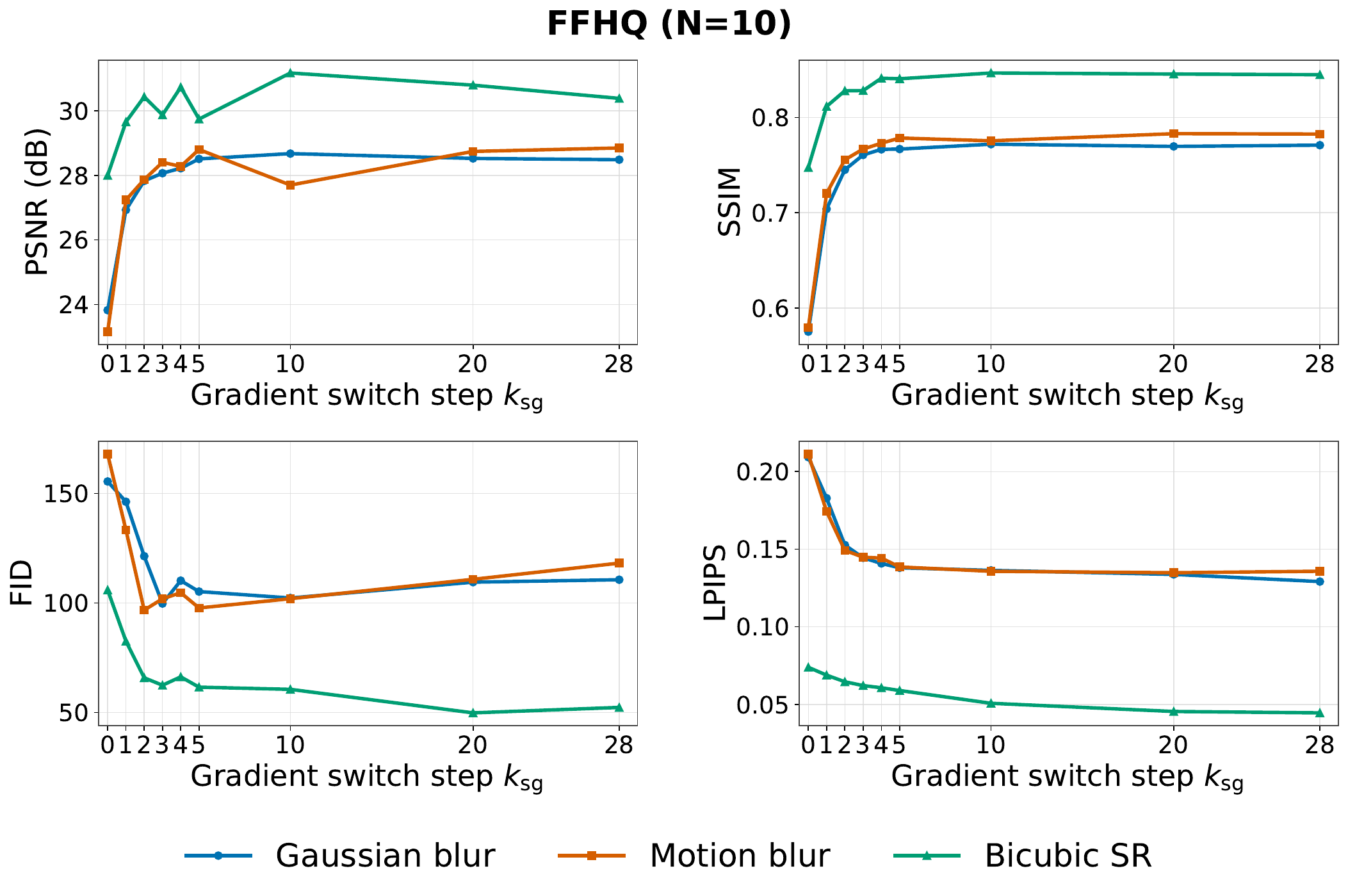}
    \caption{Ablation on gradient switch step $k_\mathrm{sg}$ on 10 FFHQ images.}
    \label{fig:image-ablation}
\end{figure}

As shown in~\cref{fig:image-ablation}, full gradients through the velocity field (\(k_{\mathrm{sg}}=0\)) yield weaker results across the evaluated metrics.
  Introducing a suitable initial stop-gradient window (\(0<k_{\mathrm{sg}}<28\)) improves performance, while \(k_{\mathrm{sg}}=28\)—which
  stops velocity gradients throughout and uses decoder-Jacobian-only feedback—also outperforms the full-gradient setting. The best
  intermediate switch point varies by task and metric.

\newpage
\subsection{Frequency Analysis}
\label{app:image-frequency}

\begin{figure}[!ht]
    \centering
    \includegraphics[width=0.98\linewidth]{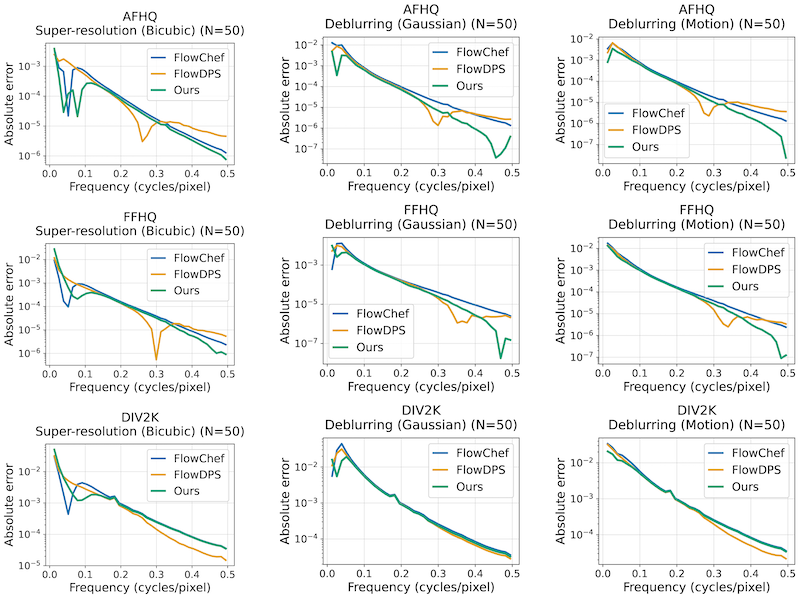}
    \caption{Spectral-density comparisons across datasets and inverse-problem tasks for $28$ NFEs.}
    \label{fig:image-psd-NFE28}
\end{figure}

The latent-space formulation uses the decoder-propagated constraint gradient by default and optionally restores the velocity-Jacobian contribution at later sampling steps when finer correction is required. To complement the pixel-space and perceptual metrics, \cref{fig:image-psd-NFE28} compares the spectral density of the reconstructions across datasets and inverse problems. This view helps distinguish improvements in spatial-frequency recovery from changes that are visible only through aggregate reconstruction metrics.

\clearpage
\subsection{Qualitative Results}
\label{app:image-vis}

\begin{figure*}[!ht]
\centering
\includegraphics[width=0.85\textwidth]{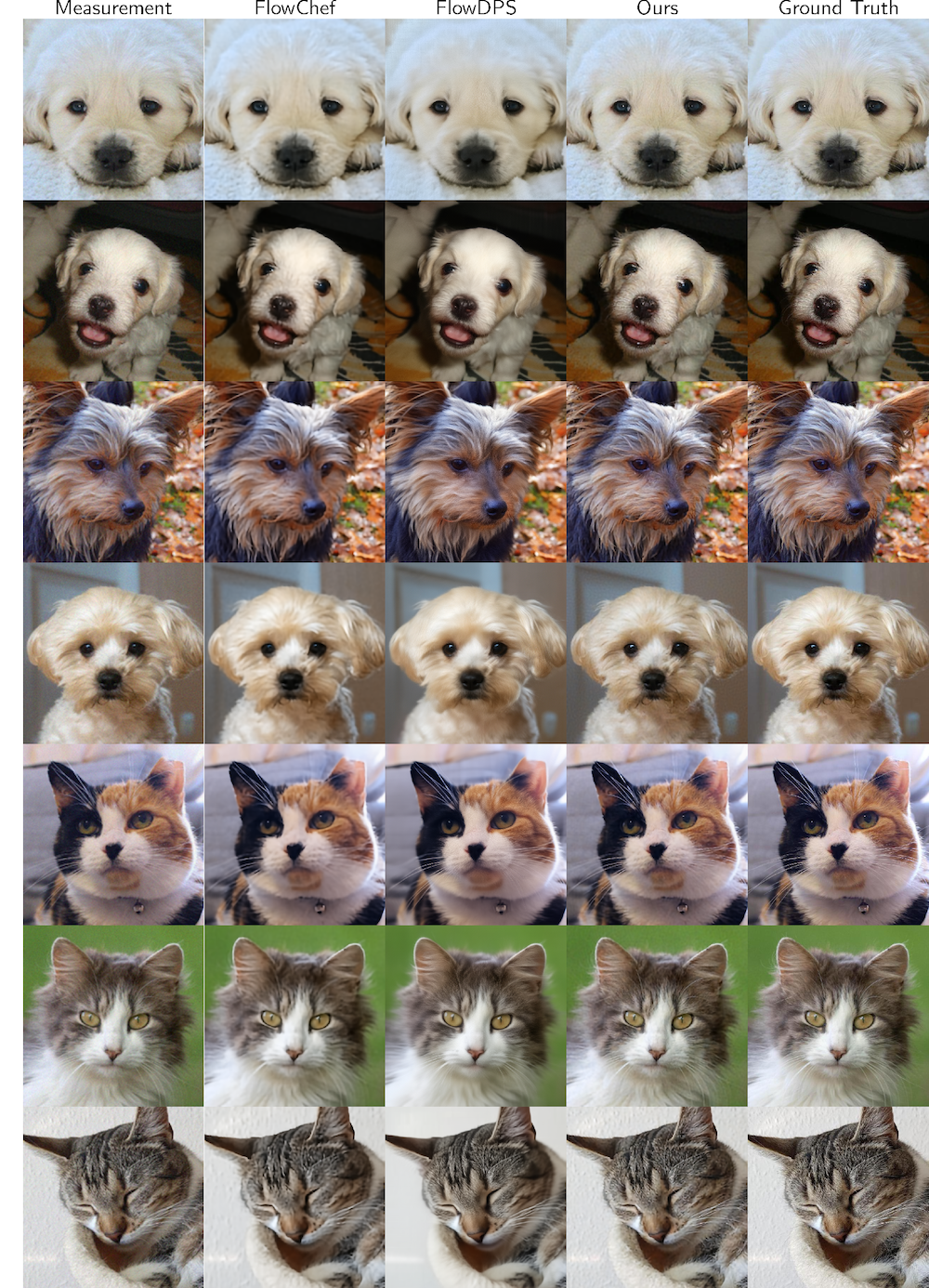}
\caption{AFHQ bicubic super-resolution using Stable Diffusion 3.5-Medium for 28 NFEs.}
\label{fig:image-afhq_sr}
\end{figure*}

\begin{figure*}[!ht]
\centering
\includegraphics[width=0.85\textwidth]{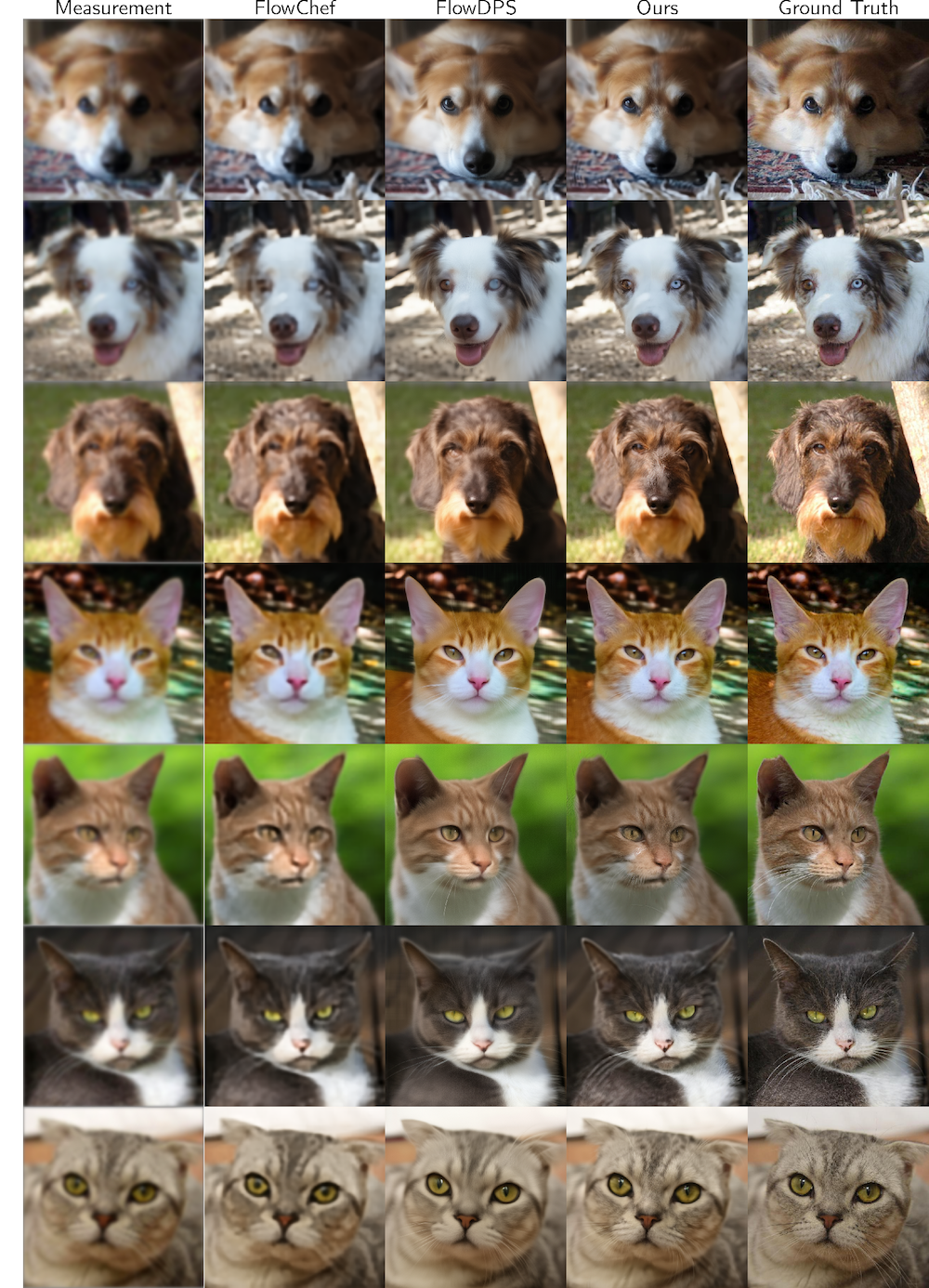}
\caption{AFHQ Gaussian deblurring using Stable Diffusion 3.5-Medium for 28 NFEs.}
\label{fig:image-afhq_gauss}
\end{figure*}

\begin{figure*}[!ht]
\centering
\includegraphics[width=0.85\textwidth]{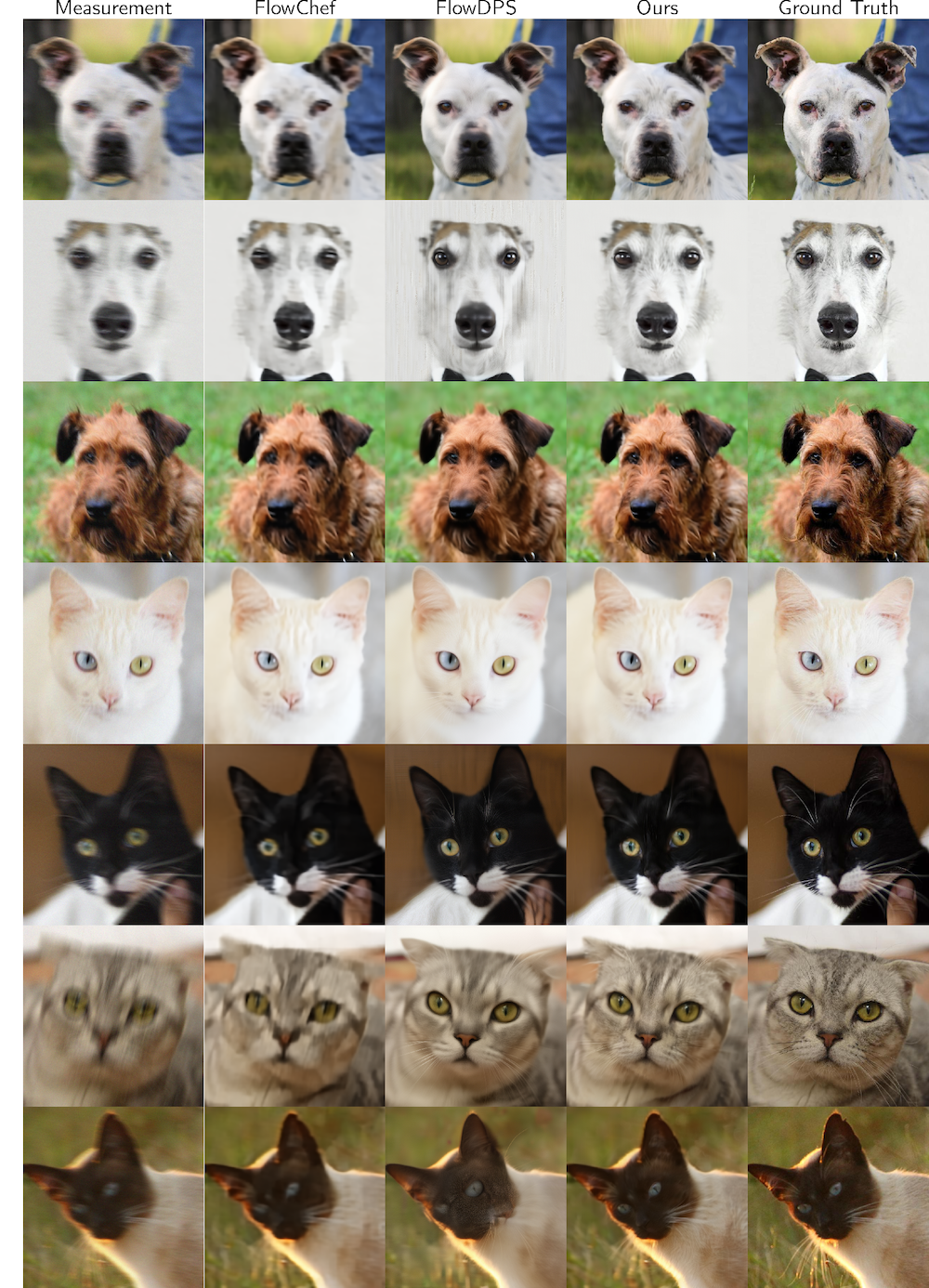}
\caption{AFHQ motion deblurring using Stable Diffusion 3.5-Medium for 28 NFEs.}
\label{fig:image-afhq_motion}
\end{figure*}

\begin{figure*}[!ht]
\centering
\includegraphics[width=0.85\textwidth]{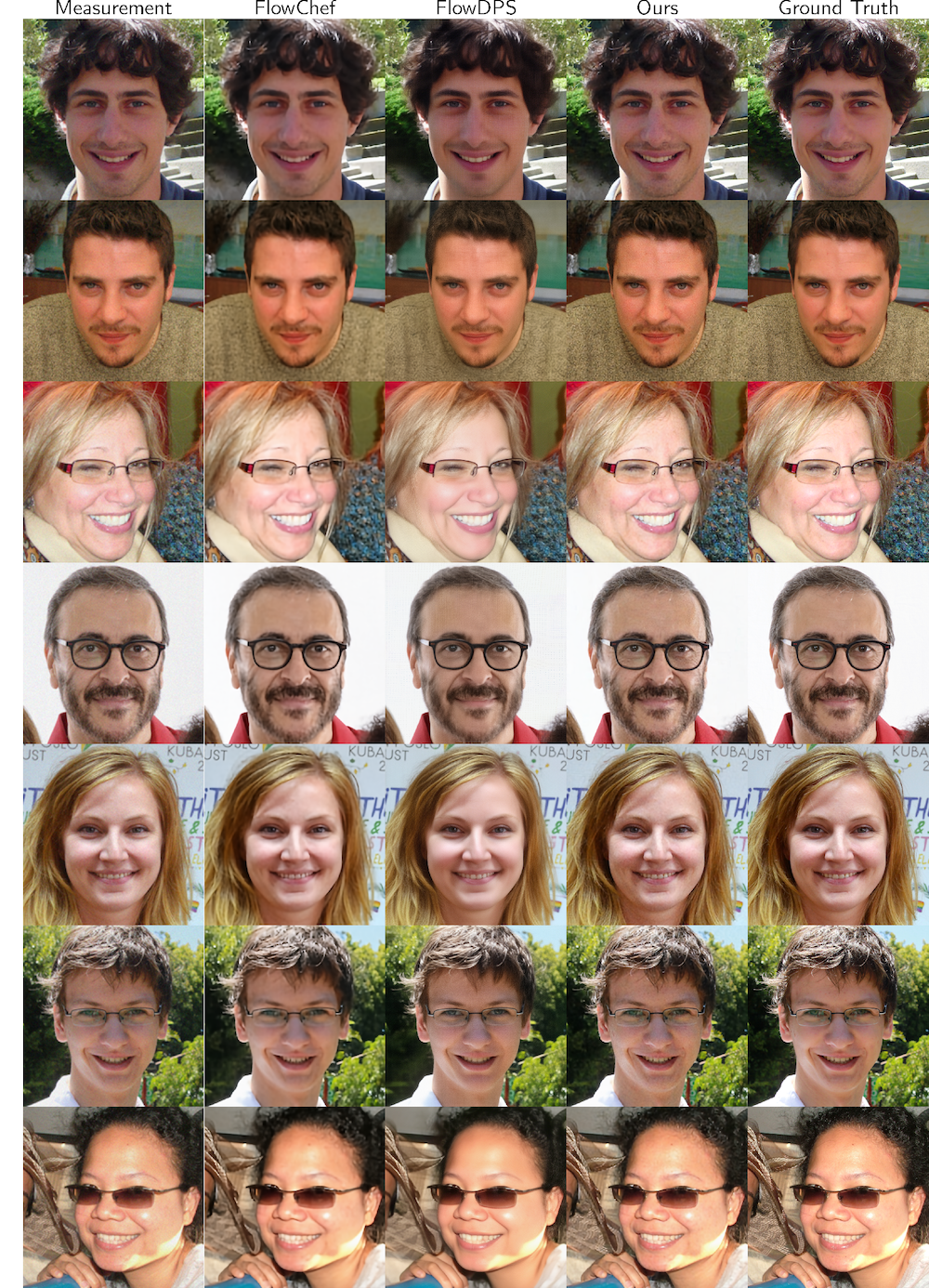}
\caption{FFHQ bicubic super-resolution using Stable Diffusion 3.5-Medium for 28 NFEs.}
\label{fig:image-ffhq_sr}
\end{figure*}

\begin{figure*}[!ht]
\centering
\includegraphics[width=0.85\textwidth]{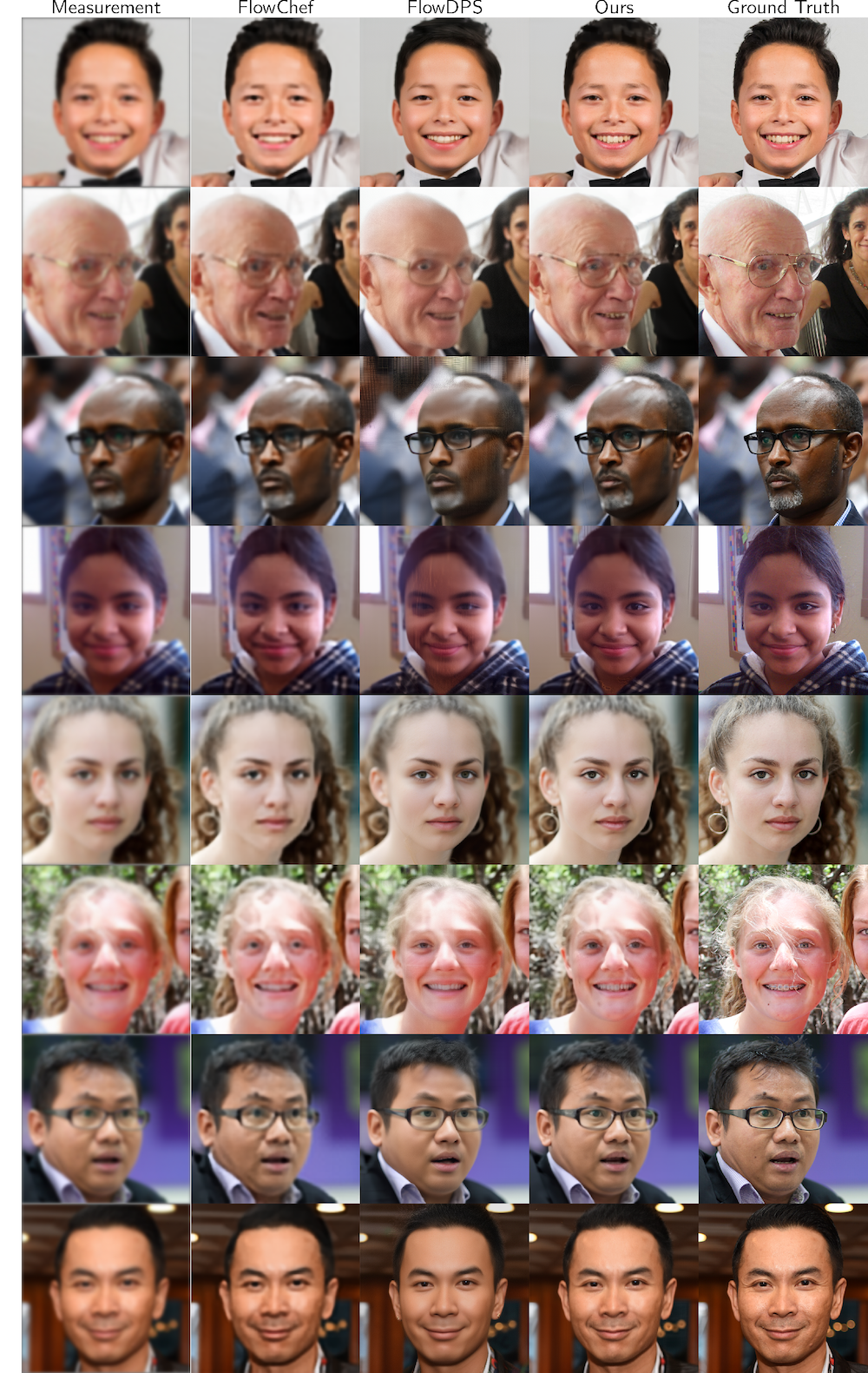}
\caption{FFHQ Gaussian deblurring using Stable Diffusion 3.5-Medium for 28 NFEs.}
\label{fig:image-ffhq_gauss}
\end{figure*}

\begin{figure*}[!ht]
\centering
\includegraphics[width=0.85\textwidth]{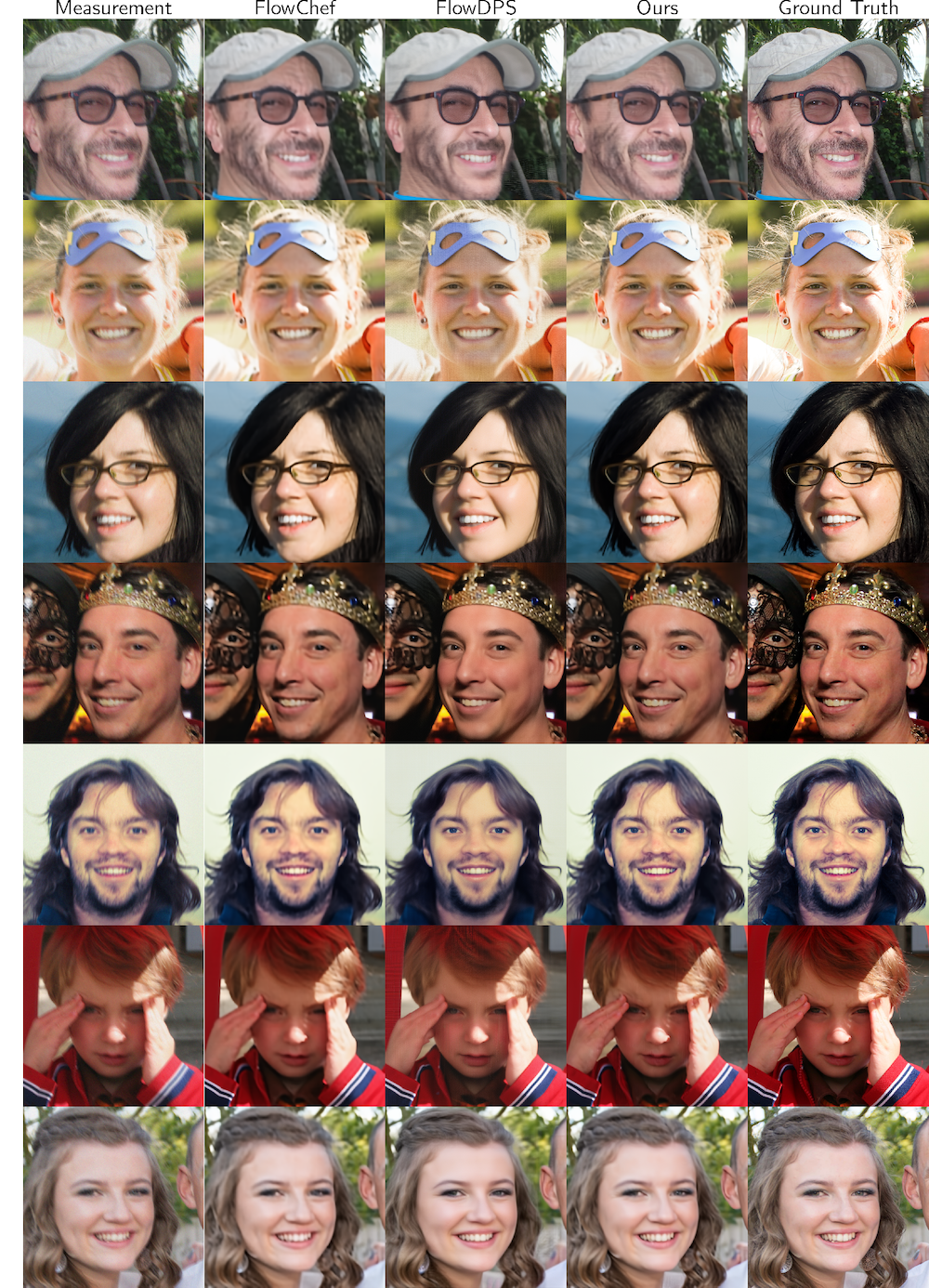}
\caption{FFHQ motion deblurring using Stable Diffusion 3.5-Medium for 28 NFEs.}
\label{fig:image-ffhq_motion}
\end{figure*}

\begin{figure*}[!ht]
\centering
\includegraphics[width=0.85\textwidth]{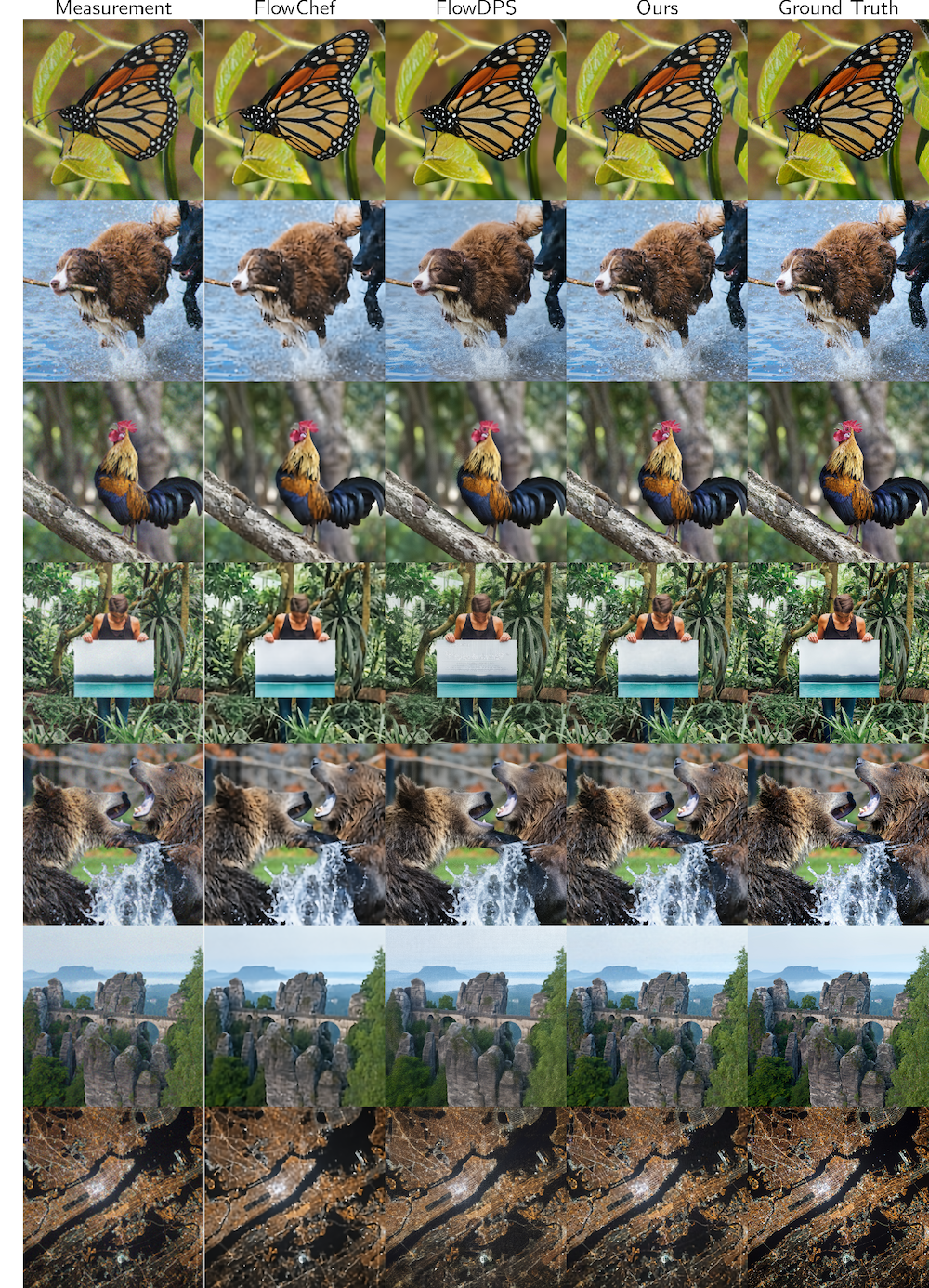}
\caption{DIV2K bicubic super-resolution using Stable Diffusion 3.5-Medium for 28 NFEs.}
\label{fig:image-div2k_sr}
\end{figure*}

\begin{figure*}[!ht]
\centering
\includegraphics[width=0.85\textwidth]{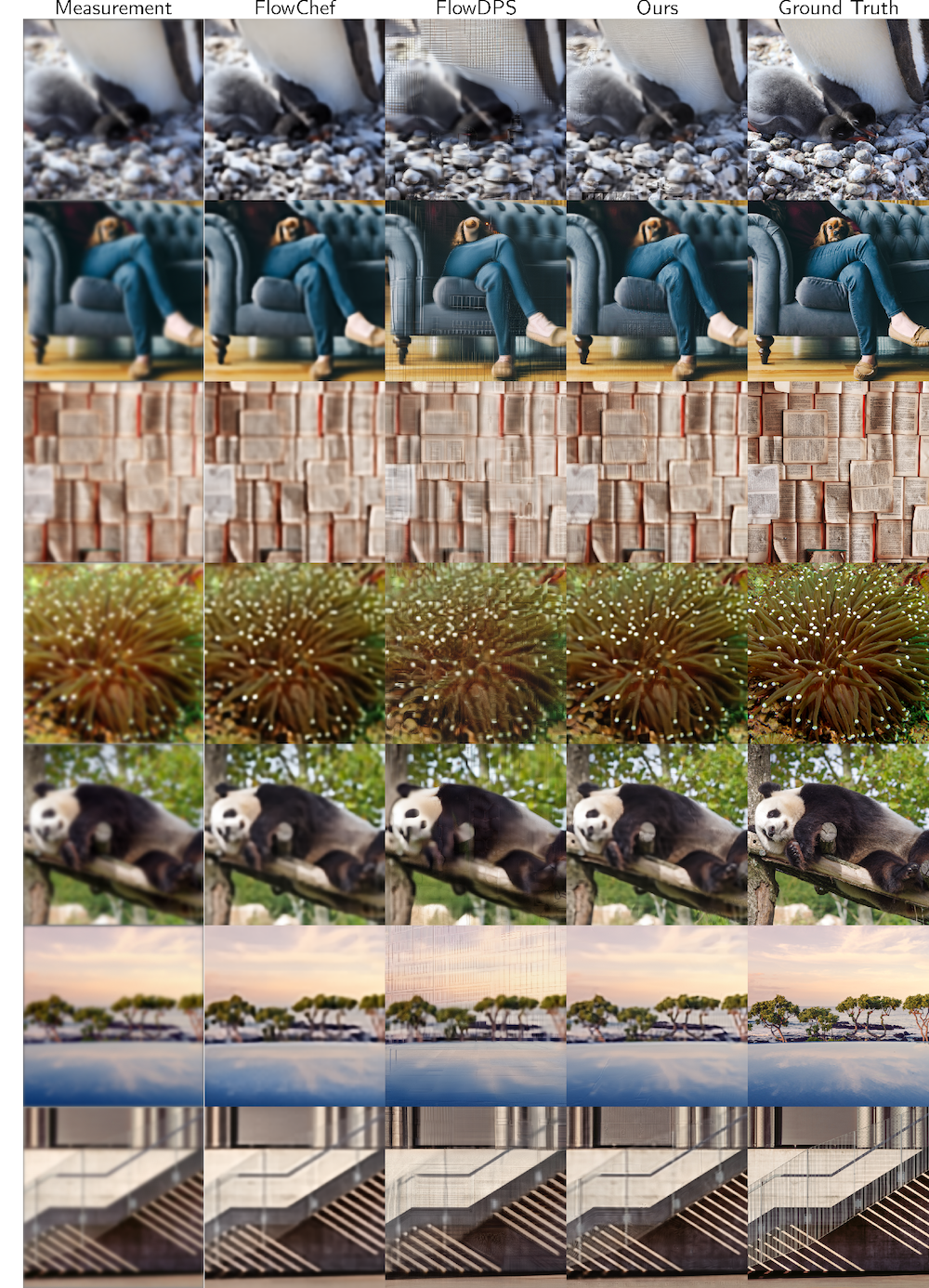}
\caption{DIV2K Gaussian deblurring using Stable Diffusion 3.5-Medium for 28 NFEs.}
\label{fig:image-div2k_gauss}
\end{figure*}

\begin{figure*}[!ht]
\centering
\includegraphics[width=0.85\textwidth]{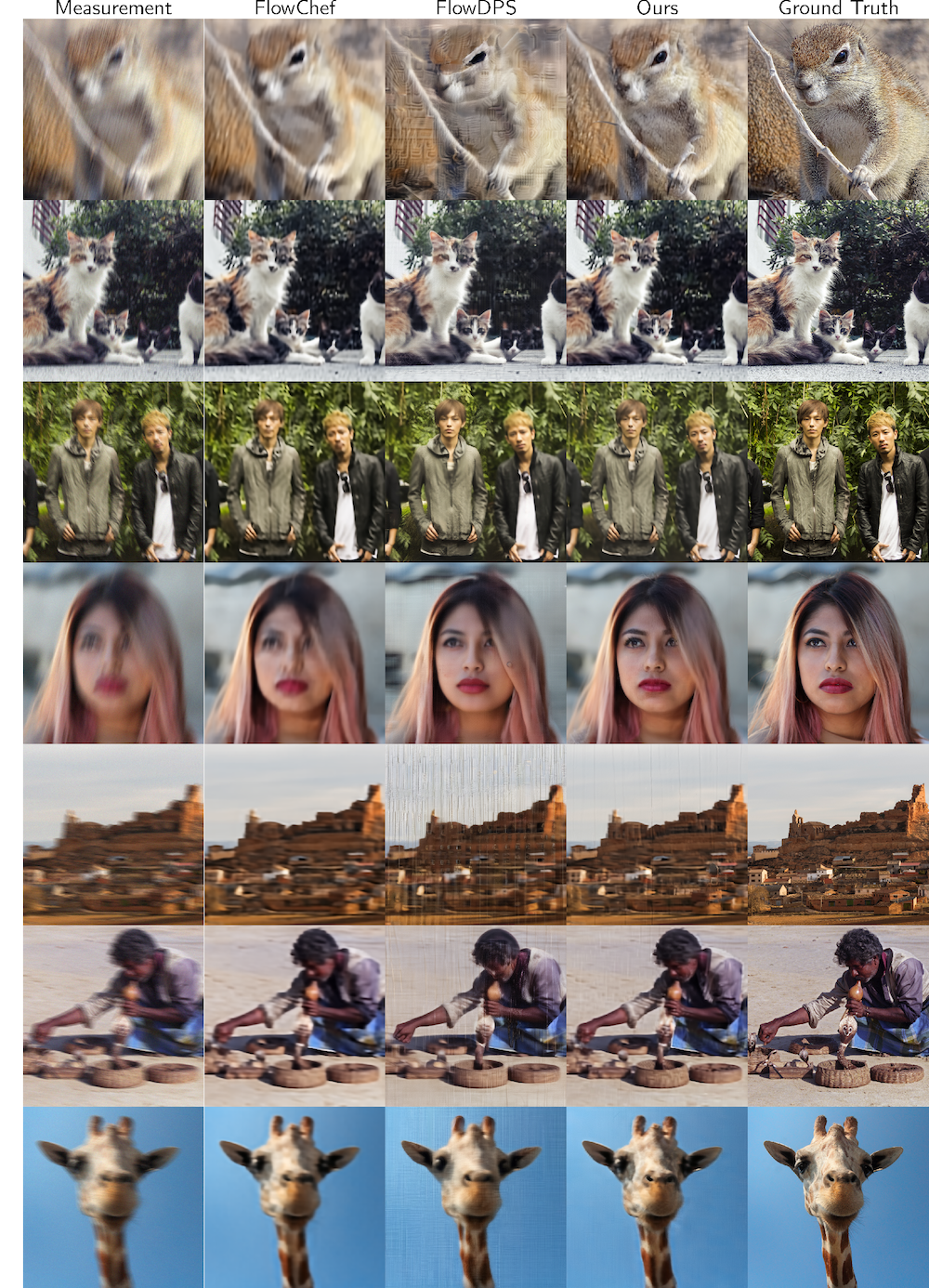}
\caption{DIV2K motion deblurring using Stable Diffusion 3.5-Medium for 28 NFEs.}
\label{fig:image-div2k_motion}
\end{figure*}

The following figures provide additional qualitative comparisons across all datasets and inverse problems. \Crefrange{fig:image-afhq_sr}{fig:image-afhq_motion} shows the AFHQ results, \crefrange{fig:image-ffhq_sr}{fig:image-ffhq_motion} the FFHQ results, and \crefrange{fig:image-div2k_sr}{fig:image-div2k_motion} the DIV2K results. Across these examples, the reconstructions complement the aggregate metrics by showing how the different guidance methods recover instance-specific structure while satisfying the observation constraint.

\end{document}